\documentclass[10pt,journal,compsoc]{IEEEtran}
\ifCLASSOPTIONcompsoc
  \usepackage[nocompress]{cite}
\else
  \usepackage{cite}
\fi
\ifCLASSINFOpdf
  \usepackage[pdftex]{graphicx}
\else
\fi
\usepackage[cmex10]{amsmath}
\usepackage{algorithmicx}
\ifCLASSOPTIONcompsoc
  \usepackage[caption=false,font=footnotesize,labelfont=sf,textfont=sf]{subfig}
\else
  \usepackage[caption=false,font=footnotesize]{subfig}
\fi
\usepackage{fixltx2e}
\usepackage{mathrsfs}
\usepackage{bm}
\usepackage{amssymb}
\usepackage{booktabs}
\usepackage{algpseudocode}

\usepackage[pagebackref=true,breaklinks=true,letterpaper=true,colorlinks,bookmarks=false]{hyperref}

\usepackage{multirow}

\newcommand{\ft}{\mathscr{F}}
\newcommand{\ftinv}{\mathscr{F}^{-1}}
\newcommand{\tp}{^\text{T}}
\newcommand{\ctp}{^*}
\newcommand{\reals}{\mathbb{R}}
\newcommand{\complexes}{\mathbb{C}}

\newsavebox{\ieeealgbox}
\newenvironment{boxedalgorithmic}
  {\begin{lrbox}{\ieeealgbox}
   \begin{minipage}{\dimexpr\columnwidth-2\fboxsep-2\fboxrule}
   \begin{algorithmic}[1]}
  {\end{algorithmic}
   \end{minipage}
   \end{lrbox}\noindent\fbox{\usebox{\ieeealgbox}}}

\algrenewcommand{\algorithmicrequire}{\textbf{Input:}}
\algrenewcommand{\algorithmicensure}{\textbf{Output:}}

\newcounter{algcount}
\newcommand{\algcaption}[1]{
	\newcounter{figcount}
	\setcounter{figcount}{\value{figure}}
	\setcounter{figure}{\value{algcount}}
	\renewcommand{\figurename}{Algorithm}
	\caption{#1}
	\setcounter{algcount}{\value{figure}}
	\setcounter{figure}{\value{figcount}}
	\vspace{3mm}}

\begin{document}
%
\title{Discriminative Scale Space Tracking}
%
%
%
%

\author{Martin~Danelljan,~\IEEEmembership{Student~Member,~IEEE,}
        Gustav~H\"ager,~\IEEEmembership{Student~Member,~IEEE,}\\
        Fahad~Shahbaz~Khan,~\IEEEmembership{Member,~IEEE,}
        and~Michael~Felsberg,~\IEEEmembership{Senior~Member,~IEEE}
\IEEEcompsocitemizethanks{\IEEEcompsocthanksitem All authors are with the Department of Electrical Engineering, Link\"oping University, SE-581 83 Link\"oping, Sweden}
}

\IEEEtitleabstractindextext{%
\begin{abstract}
Accurate scale estimation of a target is a challenging research problem in visual object tracking. Most state-of-the-art methods employ an exhaustive scale search to estimate the target size. The exhaustive search strategy is computationally expensive and struggles when encountered with large scale variations. This paper investigates the problem of accurate and robust scale estimation in a tracking-by-detection framework. We propose a novel scale adaptive tracking approach by learning separate discriminative correlation filters for translation and scale estimation. The explicit scale filter is learned online using the target appearance sampled at a set of different scales. Contrary to standard approaches, our method directly learns the appearance change induced by variations in the target scale. Additionally, we investigate strategies to reduce the computational cost of our approach.

Extensive experiments are performed on the OTB and the VOT2014 datasets. Compared to the standard exhaustive scale search, our approach achieves a gain of $2.5 \%$ in average overlap precision on the OTB dataset. Additionally, our method is computationally efficient, operating at a $50 \%$ higher frame rate compared to the exhaustive scale search. Our method obtains the top rank in performance by outperforming 19 state-of-the-art trackers on OTB and 37 state-of-the-art trackers on VOT2014. 

\end{abstract}

}

\maketitle

\IEEEdisplaynontitleabstractindextext

%
\IEEEpeerreviewmaketitle

\IEEEraisesectionheading{\section{Introduction}\label{sec:introduction}}
\IEEEPARstart{R}{obust} visual object tracking is an open research problem in computer vision with many applications in areas such as robotics, surveillance and automation. 
In generic visual tracking, only the initial location of a target is known. The task is then to estimate the trajectory of the target throughout the sequence. The problem is challenging due several factors, such as occlusions, appearance variations, motion blur, fast motion, and scale variations. 

Existing tracking approaches learn an appearance model of the target using either discriminative \cite{Henriques12d,DanelljanCVPR14,Torr11b} or generative \cite{Shengfeng13b,Jia12d,Laura12d} methods. The appearance model is then employed for estimating the target state in a new frame. In the standard case, the state includes the horizontal and vertical location of the target in the image. In many applications, such as robotics and surveillance, it is also important to estimate the target size in the image. Variations in the size of the target occur due to motion along the camera axis or changes in the target appearance. Accurate estimation of scale variations is a challenging problem and  further complicated by the presence of other factors such as occlusions, fast motion, and illumination variations.

A straightforward approach for incorporating scale estimation in a tracking framework is to evaluate the appearance model at multiple resolutions by performing an exhaustive scale search. However, this brute-force search strategy is computationally demanding. In real-time applications, computational efficiency is a crucial factor. Therefore, an ideal tracking approach should be robust with respect to scale variations while operating at real-time. In this work, we investigate the problem of accurate scale adaptive visual tracking with an emphasis on real-time performance.

Recently, discriminative correlation filter (DCF) based visual trackers \cite{MOSSE2010,Henriques12d,DanelljanCVPR14} have shown to provide excellent performance. Moreover, these trackers have the advantage of being computationally efficient, thereby making them especially suitable for a variety of real-time applications. The success of DCF-based trackers is evident from the results of the Visual Object Tracking (VOT) 2014 Challenge \cite{VOT2014}, where all the top three trackers are based on correlation filters. Related DCF based methods \cite{DanelljanCVPR14,henriquesARXIV14} have also shown competitive performance on the OTB dataset \cite{Wu13g}, while operating at over 100 frames per second. The DCF based methods work by learning an optimal correlation filter used to locate the target in the next frame. The significant gain in speed is obtained by exploiting the fast Fourier transform (FFT) at both learning and detection stages. Most methods that employ DCFs for tracking mainly focus on the problem of translation estimation. Instead, we investigate DCF based methods for real-time scale adaptive visual tracking.

\begin{figure*}[htbp]
  \begin{center}
          \newcommand{\wid}{0.25\textwidth}
          \includegraphics[width=\wid]{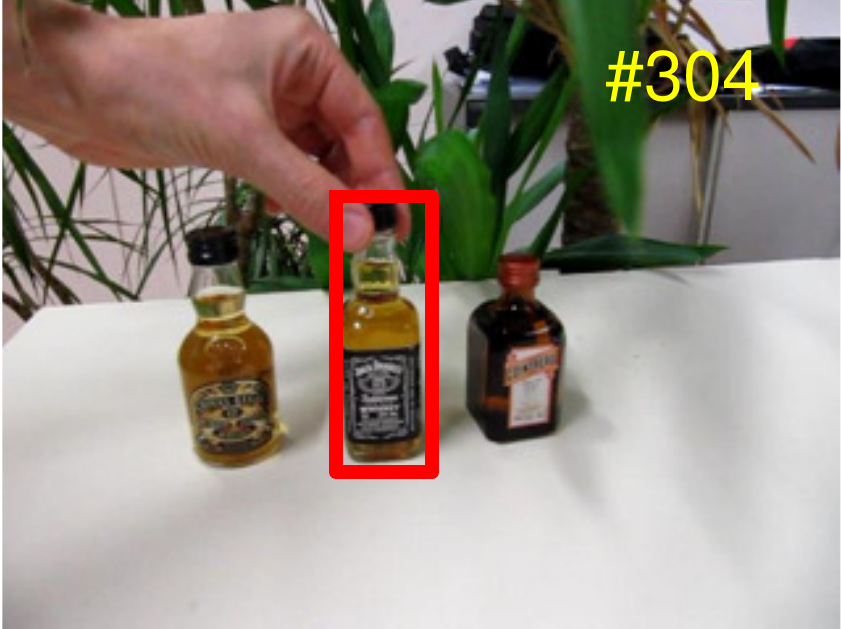}%
          \includegraphics[width=\wid]{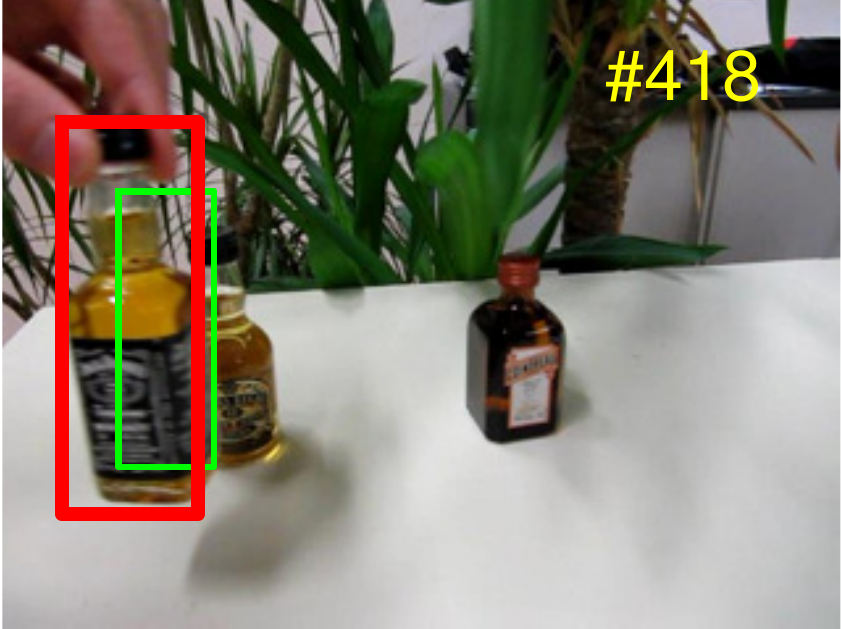}%
          \includegraphics[width=\wid]{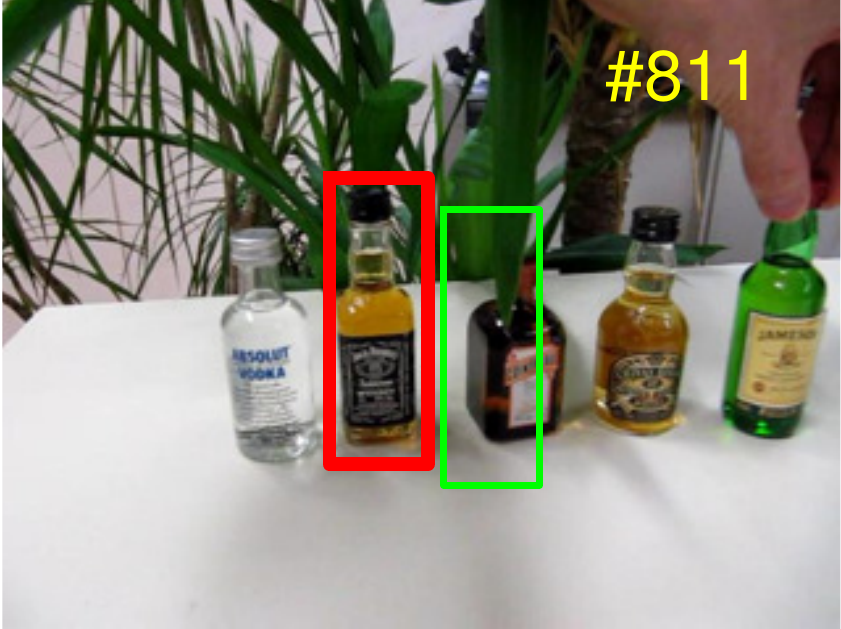}%
          \includegraphics[width=\wid]{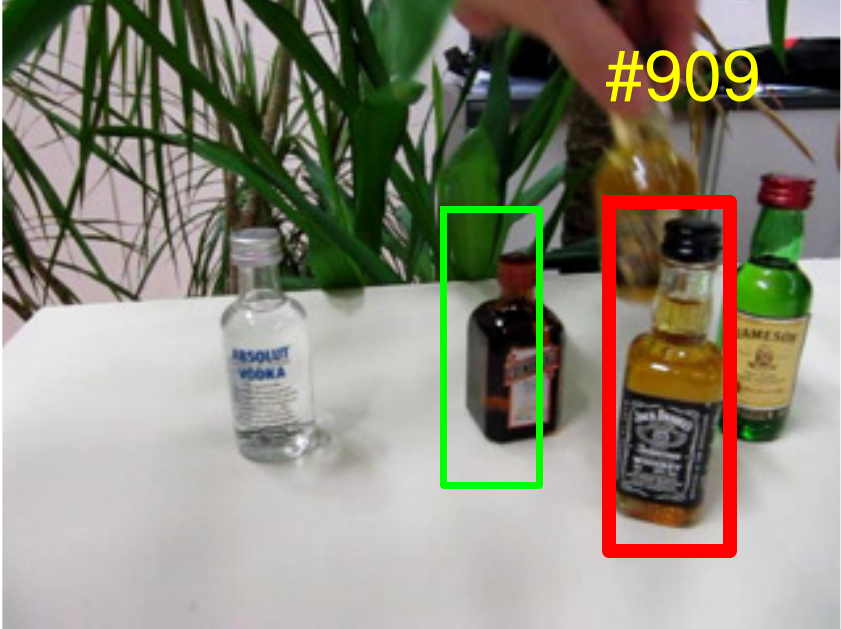}%
		  \vspace{1mm}
          \includegraphics[width=\wid]{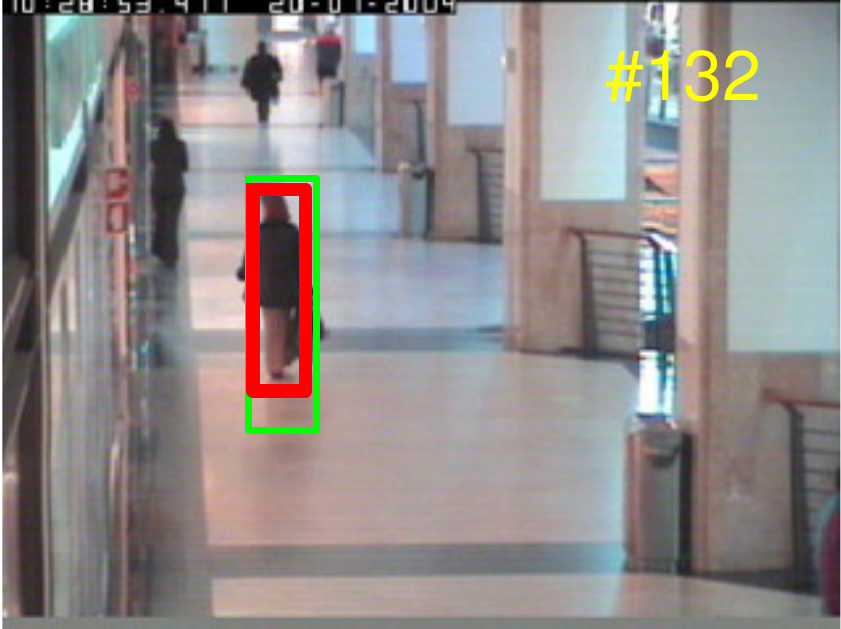}%
          \includegraphics[width=\wid]{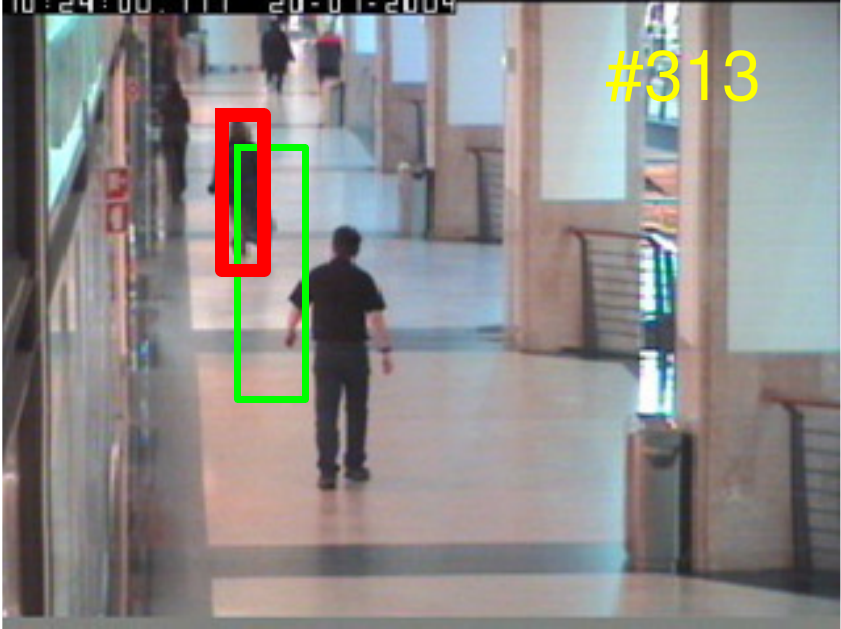}%
          \includegraphics[width=\wid]{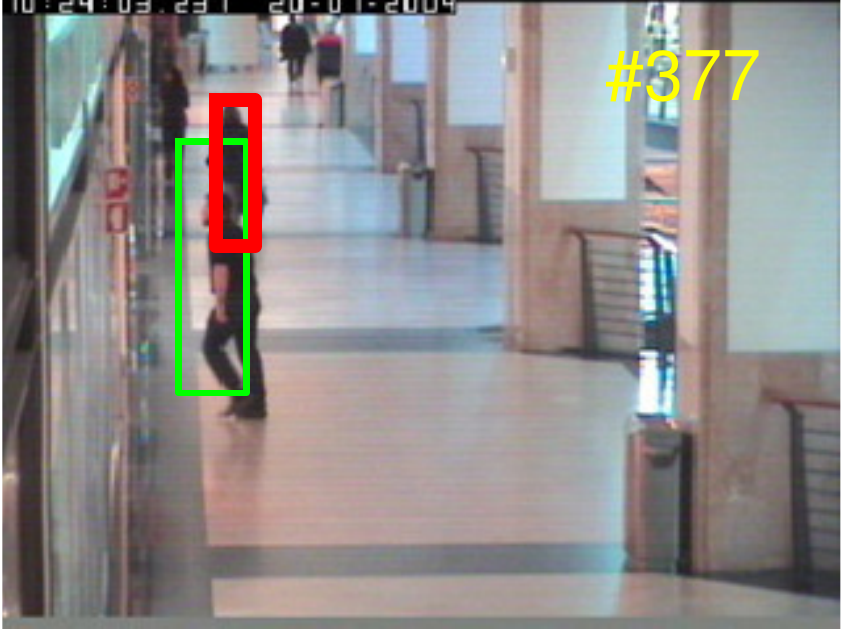}%
          \includegraphics[width=\wid]{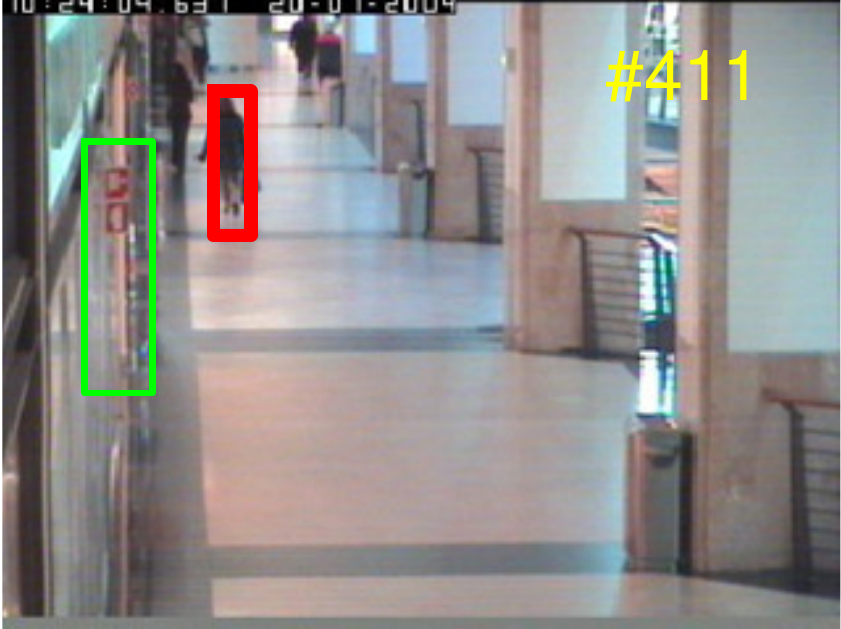}%
		  \vspace{1mm}
          \includegraphics[width=\wid]{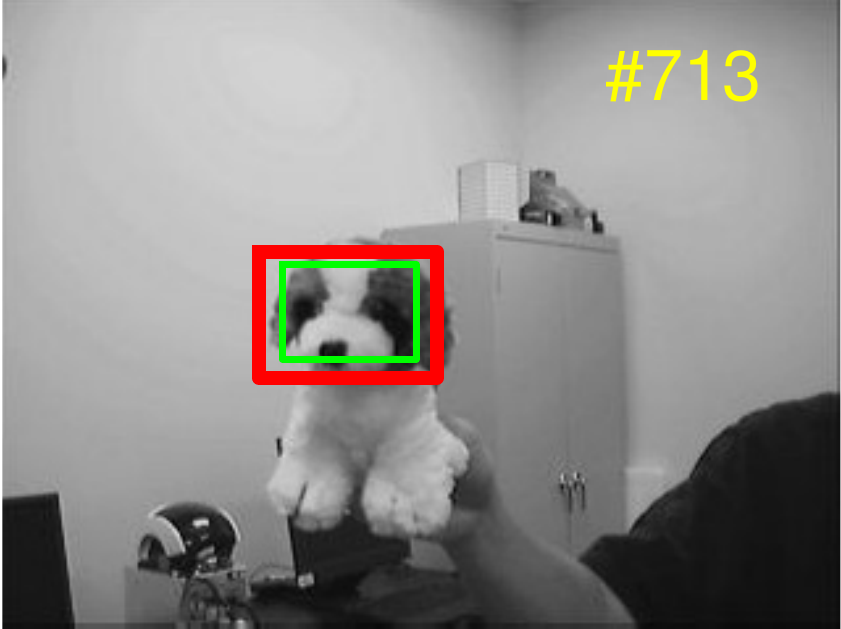}%
          \includegraphics[width=\wid]{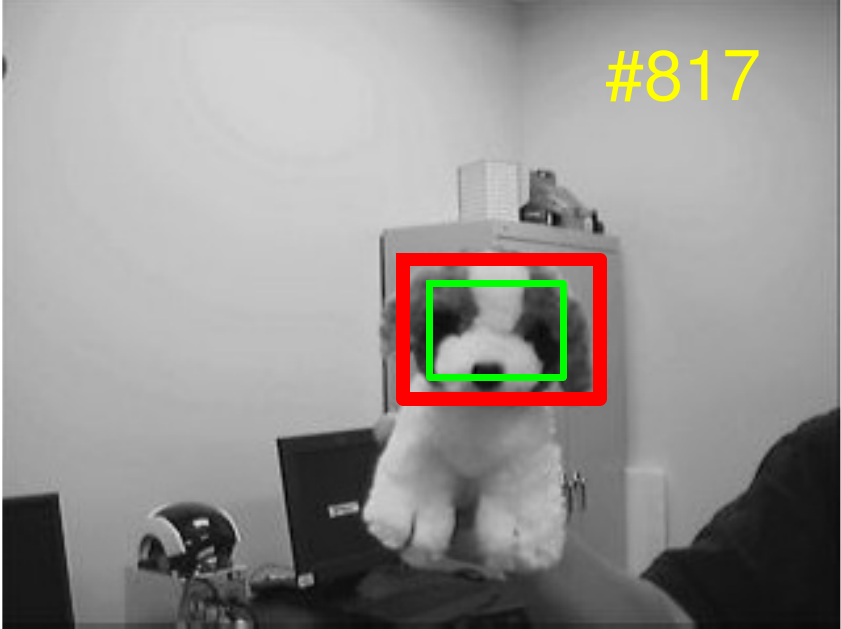}%
          \includegraphics[width=\wid]{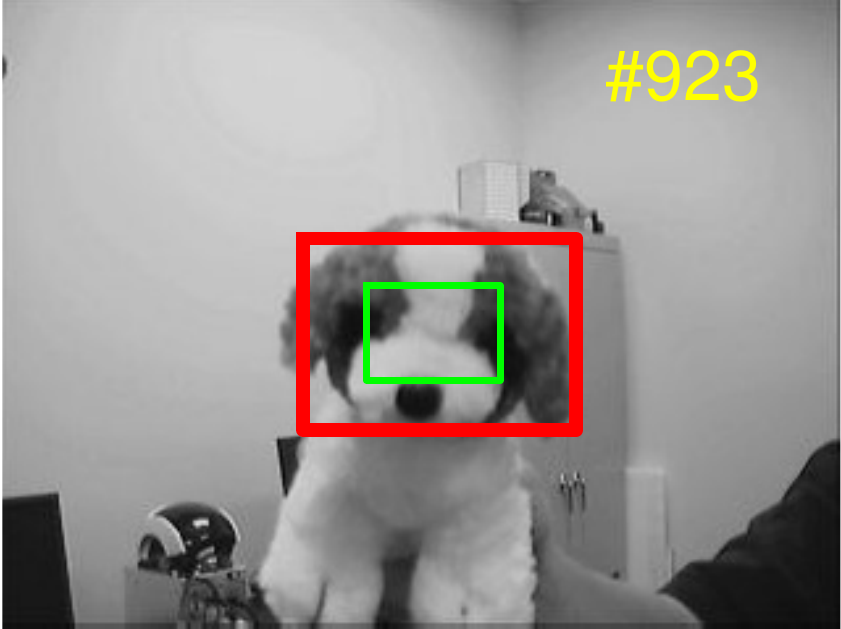}%
          \includegraphics[width=\wid]{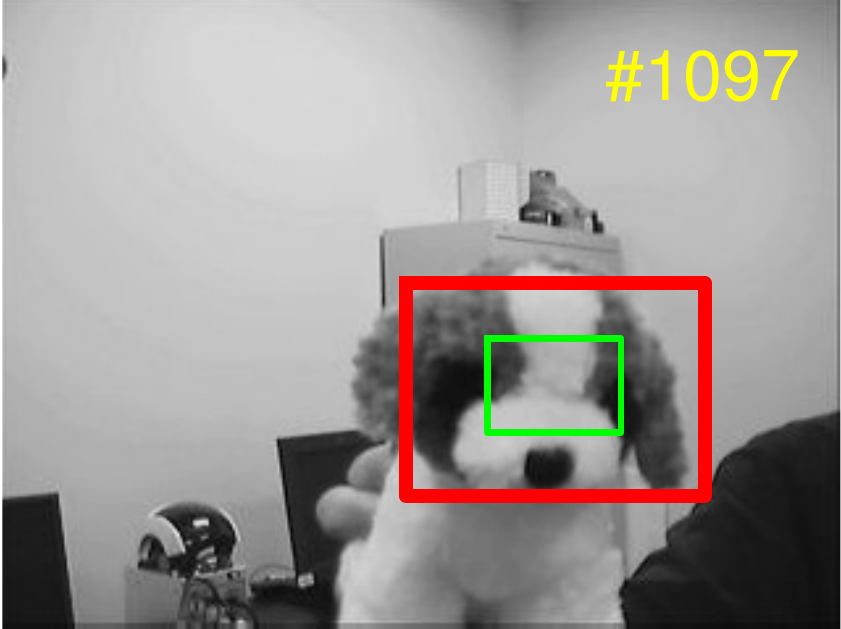}%
  \end{center}
  \caption{Comparison of our discriminative scale space tracker (in red) with the standard DCF based translation tracker (in green). Example frames are shown from the \emph{liqour} (top row), \emph{walking} (middle row) and \emph{dog} (bottom row) sequences. The standard DCF tracker is not able to handle the scale changes in these sequences. In contrast, our approach accurately estimates the target size, thereby significantly increasing both the accuracy and robustness of the tracker.}
  \label{fig:scale_problems}
\end{figure*}

To incorporate scale estimation in a DCF based tracking framework, two different exhaustive scale search strategies are considered. A first approach, \emph{joint scale space filter}, works by constructing a 3-dimensional correlation filter for joint estimation of translation and scale. A second strategy, \emph{multi-resolution translation filter}, applies a standard 2-dimensional translation correlation filter at multiple resolutions. However, both these approaches are computationally demanding and not suitable for real-world tracking applications.
In this work, we propose an alternative, discriminative approach for scale adaptive visual tracking.

We propose the \emph{discriminative scale space tracker} (DSST), that learns separate correlation filters for explicit translation and scale estimation. The scale filter is learned using samples of the target at a set of different scales. Given a new frame, we first estimate the target translation using a standard translation filter. Afterwards,  we apply the learned scale filter at the target location to obtain an accurate estimate of the target size.
Different from the aforementioned exhaustive scale search strategies, our method explicitly learns the appearance change induced by variations in the target size while reducing the search space.
We further investigate strategies to reduce the computational cost of our DSST approach without sacrificing its robustness and accuracy. The reduced cost allows us to increase the target search area of the tracker for improved robustness. The resulting fast DSST tracker (fDSST) achieves significantly improved tracking performance, while providing a twofold gain in speed compared to the DSST method.  

To validate the performance of our approach, a comprehensive evaluation is performed on the full Online Tracking Benchmark (OTB) dataset \cite{Wu13g}, containing 50 videos. We also evaluate the performance our method on the Visual Object Tracking (VOT) 2014 challenge \cite{VOT2014}. This dataset contains 25 representative videos, that were selected for the VOT 2014 challenge from an initial pool of 394 videos \cite{VOT2014}. Both quantitative and qualitative experiments are performed. Our DSST approach improves the baseline DCF exhaustive scale search methods both in terms of accuracy and speed. We further show that our tracker outperforms 19 compared state-of-the-art methods on the OTB dataset. Finally, our approach is shown to obtain the top rank on the VOT 2014 dataset by outperforming 37 state-of-the-art trackers. It is worth mentioning that our proposed approach is generic and can be incorporated in any tracking method with no scale estimation component. Fig.~\ref{fig:scale_problems} shows a comparison of our approach with the standard DCF based translation tracker. The translation tracker struggles in the presence of scale variations, while our approach accurately estimates the target size and thereby significantly improves the robustness and accuracy of the tracker.

The rest of this paper is organized as follows. Section~\ref{sec:related} gives an overview of the prior work most relevant to this work. In section~\ref{sec:MDCF} we introduce multi-channel discriminative correlation filters for visual tracking. Section~\ref{sec:appraches} discusses the DCF-based translation tracker and strategies for extending the DCF framework to include scale estimation. The proposed discriminative scale space tracker (DSST) and the fast DSST are described in section~\ref{sec:our_approach}. In section~\ref{sec:experiments} we provide description and results of the performed experiments. Conclusions are finally given in section~\ref{sec:conclusion}.

\section{Related Work}
\label{sec:related}
Visual object tracking is a fundamental computer vision problem. The objective is to estimate the trajectory of a target in an image sequence. In generic visual tracking, the target can be any object and is defined solely by its initial location. The initial target location is hence the only information given to the tracker.
Due to the general nature of this problem, it is applicable in a large variety of computer vision tasks. Popular applications of generic tracking include robotics \cite{DanelljanECCVW14}, surveillance \cite{ProkajCVPR14} and road scene understanding \cite{Geiger2014PAMI}.

Typically, visual tracking methods work by constructing a target appearance model from the observed image information. This is achieved using either generative \cite{Laura12d,Bao12d,Kwon11d,Liu11d} or discriminative \cite{Torr11b,Zhang12c,Babenko09b,Henriques12d} approaches. Generative appearance models aim at describing the target appearance using e.g.\ statistical models or templates. The discriminative approaches instead employ machine learning techniques to differentiate between the target appearance and the surrounding background. Examples  of employed learning approaches include Support Vector Machines (SVM) \cite{Torr11b} and boosting techniques \cite{Babenko09b}. 

Recently, Discriminative Correlation Filters (DCF) have successfully been applied to visual tracking \cite{MOSSE2010,DanelljanCVPR14,henriquesARXIV14,DanelljanBMVC14}. These methods have shown to provide excellent results on benchmark tracking datasets \cite{Wu13g,VOT2014}, while operating at real-time. The correlation-based trackers learn a discriminative correlation filter for locating the target in each new frame. Bolme et al.\ \cite{MOSSE2010} trained the filter by minimizing the total squared error between the actual and the desired correlation output on a set of sample grayscale patches. By using circular correlation, the authors showed that the resulting filter can be computed efficiently using only FFTs and pointwise operations. Henriques et al.\ \cite{Henriques12d} further showed that the DCF formulation equivalently can be cast as learning a least squares regressor (ridge regression) on the set of all cyclic shifts of the involved training sample patches. This formulation was then used to introduce fast kernelized correlation filters.

Several works have recently investigated generalizations of the DCF approach \cite{MOSSE2010} for multidimensional features \cite{galoogahiICCV13,BoddetiCVPR13,henriquesICCV13}. These approaches consider learning an exact multi-channel filter given the set of training samples. However, such methods are not directly applicable to the online tracking problem due to a significant increase in the computational cost. Alternatively, approximate formulations for learning multi-channel filters have been investigated for visual tracking \cite{DanelljanCVPR14,henriquesARXIV14}. These approaches have shown to be robust while only scaling linearly with the number of feature channels. Furthermore, Danelljan et al.\ \cite{DanelljanCVPR14} introduced an adaptive feature dimensionality reduction technique to reduce the computational cost while preserving tracking performance.

The DCF based approaches have demonstrated the capability of accurate target localization in many different challenging tracking scenarios. However, the standard DCF trackers are restricted to translation estimation. This implies poor performance when encountered with significant variations in the target scale. Furthermore, the capability of accurately retrieving the target scale is beneficial in many tracking applications. Recently, a multi-resolution extension of a kernelized correlation translation filter was proposed by Li and Zhu \cite{Li2014}. However, this approach suffers from a higher computational cost, since the translation filter has to be applied at several resolutions to achieve sufficient scale accuracy. Contrary to \cite{Li2014}, we aim at directly learning the appearance changes induced by scale variations. This allows us to achieve accurate scale adaptive tracking at a significantly higher frame-rate.

This paper extends our work \cite{DanelljanBMVC14}, which is the winning method in the Visual Object Tracking (VOT) 2014 challenge \cite{VOT2014}. In this paper, we additionally perform a comprehensive analysis of DCF based approaches for scale adaptive visual tracking. Furthermore, we extend our Discriminative Scale Space Tracker (DSST) \cite{DanelljanBMVC14} by investigating strategies to reduce its computational cost. This enhancement further allows us to increase the robustness by extending the target search area, without sacrificing real-time performance. The proposed improvements result in superior tracking performance and a twofold speedup. The experiments are extended by evaluating our approach on the full (50 videos) OTB dataset and comparing with 19 state-of-the-art trackers. Finally, we also present results on the VOT 2014 dataset. 

\section{Multi-channel Discriminative Correlation Filters}
\label{sec:MDCF}
Our tracking approach is based on learning discriminative correlation filters (DCF) \cite{MOSSE2010}. 
The DCF based tracking approaches learn an optimal correlation filter for locating the target in a new frame, given a set of sample patches of the target appearance. This can equivalently be formulated as learning a classifier based on all cyclic shifts of the sample target patches \cite{Henriques12d}. The DCF approach has recently been extended to multidimensional feature representations for a number of applications, including visual tracking \cite{DanelljanCVPR14,henriquesARXIV14}, object detection \cite{henriquesICCV13,galoogahiICCV13} and object alignment \cite{BoddetiCVPR13}. In this work, we utilize multi-channel DCFs for a variety of tasks within visual tracking.

We first introduce learning a multi-channel correlation filter from a single sample $f$ of the target appearance. In the standard case  $f$ corresponds to an image patch centered around the target. This is used to learn a 2-dimensional correlation filter for estimating the target translation. Generally, the dimension of the domain of $f$ is arbitrary. Therefore, the same approach can be used to learn 1-dimensional scale estimation filters, 2-dimensional translation estimation filters and 3-dimensional joint scale and translation estimation filters. This is accomplished by only adapting the feature extraction step for each case.

The target sample $f$ consists of a $d$-dimensional feature vector $f(\bm{n}) \in \reals^d$, at each location $\bm{n}$ in a rectangular domain. In the translation case we may e.g.\ use the RGB-value at each pixel location within the patch. In general however, we can consider any grid-based feature representation. We denote feature channel $l \in \{1, \ldots, d\}$ of $f$ by $f^l$. The objective is then to learn a correlation filter $h$ consisting of one filter $h^l$ per feature channel. This is achieved by minimizing the $L^2$ error of the correlation response compared to the desired correlation output $g$,
\begin{equation}
	\label{eq:multiple_feat_cost}
	\varepsilon = \Bigg\| g - \sum_{l=1}^d h^l \star f^l \Bigg\|^2 + \lambda \sum_{l=1}^d \big\| h^l \big\|^2 .
\end{equation}
Here, the star $\star$ denotes circular correlation. The second term in \eqref{eq:multiple_feat_cost} is a regularization with a weight parameter $\lambda$. The desired correlation output $g$ is typically selected to be a Gaussian function with a parametrized standard deviation \cite{MOSSE2010}. Note that the domains of $f^l$, $h^l$ and $g$ all have the same dimension and size.

Eq.~\ref{eq:multiple_feat_cost} is a linear least squares problem. It can be solved efficiently by transforming \eqref{eq:multiple_feat_cost} to the Fourier domain using Parseval's formula. The filter that minimizes \eqref{eq:multiple_feat_cost} is given by
\begin{equation}
	\label{eq:multiple_feat_sol}
	H^l = \frac{\overline{G} F^l}{\sum_{k=1}^d \overline{F^k} F^k + \lambda} \; , \quad l = 1, \ldots, d .
\end{equation}
Here, the capital letters denote the discrete Fourier transform (DFT) of the corresponding quantities. The bar $\overline{G}$ denotes complex conjugation. The multiplications and divisions in \eqref{eq:multiple_feat_sol} are performed pointwise. A detailed derivation of \eqref{eq:multiple_feat_sol} is provided in the appendix.

Eq.~\ref{eq:multiple_feat_sol} provides the optimal filter $h$ given a single training sample $f$ of the target appearance. In practice several samples $\{f_j\}_1^t$ at different time instances, need to be considered in order to learn a robust correlation filter $h$. This can be achieved by averaging the correlation error in \eqref{eq:multiple_feat_cost} over all training samples $f_1,\ldots,f_t$. As shown by Galoogahi et al.\ \cite{galoogahiICCV13}, the resulting linear least squares problem can be block diagonalized by the DFT. The resulting $H$ can then be found by solving $N$ number of $d\times d$ linear systems, where $N$ is equal to the number of elements in the filter $h^l$. However, this results in a computational bottleneck for our online learning task. Therefore, we compute a robust approximation by utilizing the exact solution for the single training sample case \eqref{eq:multiple_feat_sol}. Inspired by the update rule derived for the single feature case ($d=1$) \cite{MOSSE2010}, we update the numerator $A^l_t$ and denominator $B_t$ of the filter $H^l_t$ with a new sample $f_t$ as
\begin{subequations}
	\label{eq:our_training}
	\begin{align}
		\label{eq:our_training_num}
		A_t^l &= (1 - \eta) A_{t-1}^l + \eta \overline{G} F_t^l \; , \quad l = 1, \ldots, d \\
		\label{eq:our_training_den}
		B_t &= (1 - \eta) B_{t-1} + \eta \sum_{k=1}^d \overline{F^k_t} F_t^k
	\end{align}
\end{subequations}
Here, the scalar $\eta$ is a learning rate parameter.

To apply the filter in a new frame $t$, a sample $z_t$ is extracted from a considered region of transformations. In the standard translation filter case, $z_t$ corresponds to an image patch centered around the predicted target location. The test sample $z_t$ is extracted similarly to the training samples $f_t$, using the same feature representation. The DFT of the correlation scores $y_t$ is computed in the Fourier domain
\begin{equation}
	\label{eq:our_detection}
	Y_t = \frac{\sum_{l=1}^d \overline{A^l_{t-1}} Z^l_t}{B_{t-1} + \lambda} .
\end{equation}
Here, $A^l_{t-1}$ and $B_{t-1}$ are the numerator and denominator of the filter updated in the previous frame. The correlation scores at the locations reflected in $z_t$ are then computed by taking the inverse DFT $y_t = \ftinv\{Y_t\}$. The estimate of the current target state is obtained by finding the maximum correlation score.

\begin{figure}[!t]
   \centering
   \includegraphics[scale=0.25]{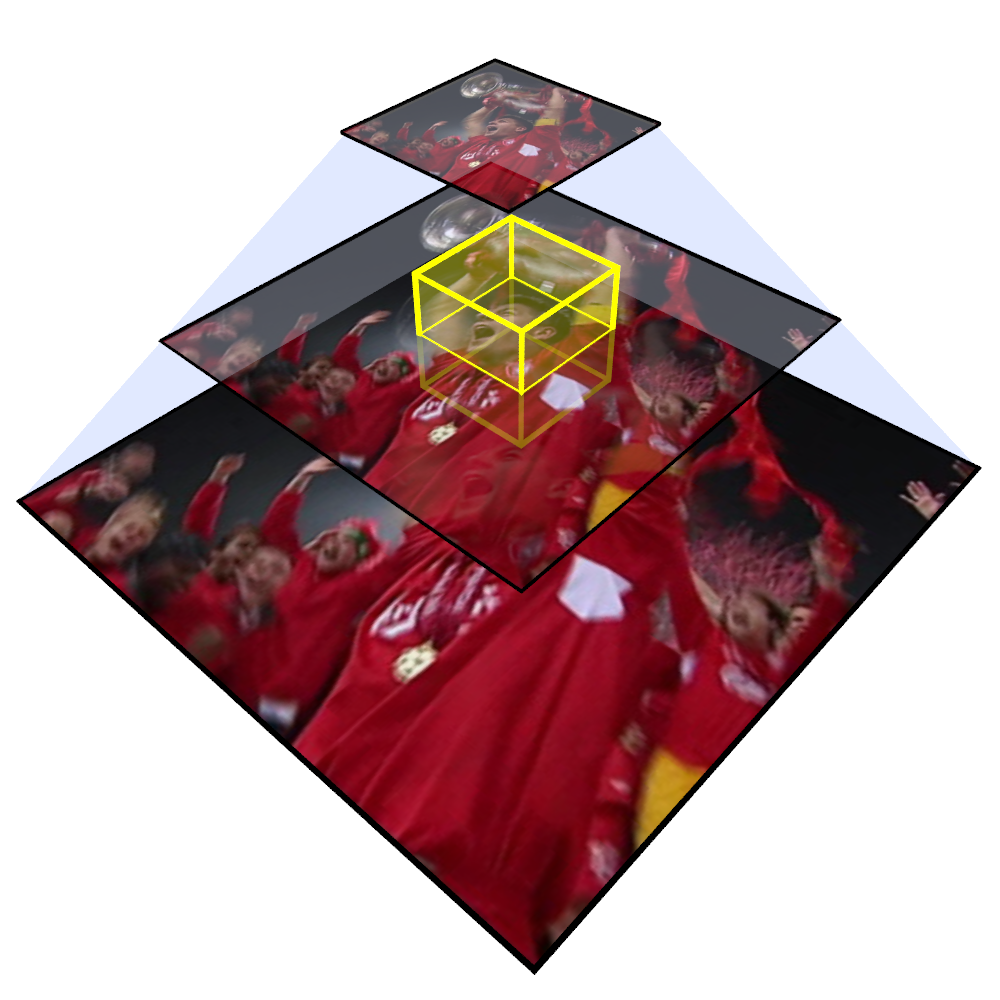}
  \caption{Visualization of the joint scale space filter approach. The training sample (yellow box) is extracted from a scale pyramid constructed around the target center.}
  \label{fig:joint_sample}
\end{figure}

\section{Correlation Filters for Translation and Scale Estimation}
\label{sec:appraches}
In this section, we investigate different approaches for translation and scale estimation in a DCF based tracking framework. We first describe the standard DCF based translation tracking approach. Afterwards, we investigate different strategies for scale estimation. 
 
\subsection{Standard DCF Tracker}
\label{sec:baseline}
\begin{figure*}[!t]
   \centering
   \subfloat[Translation filter sample.]{%
   \includegraphics[scale=0.22]{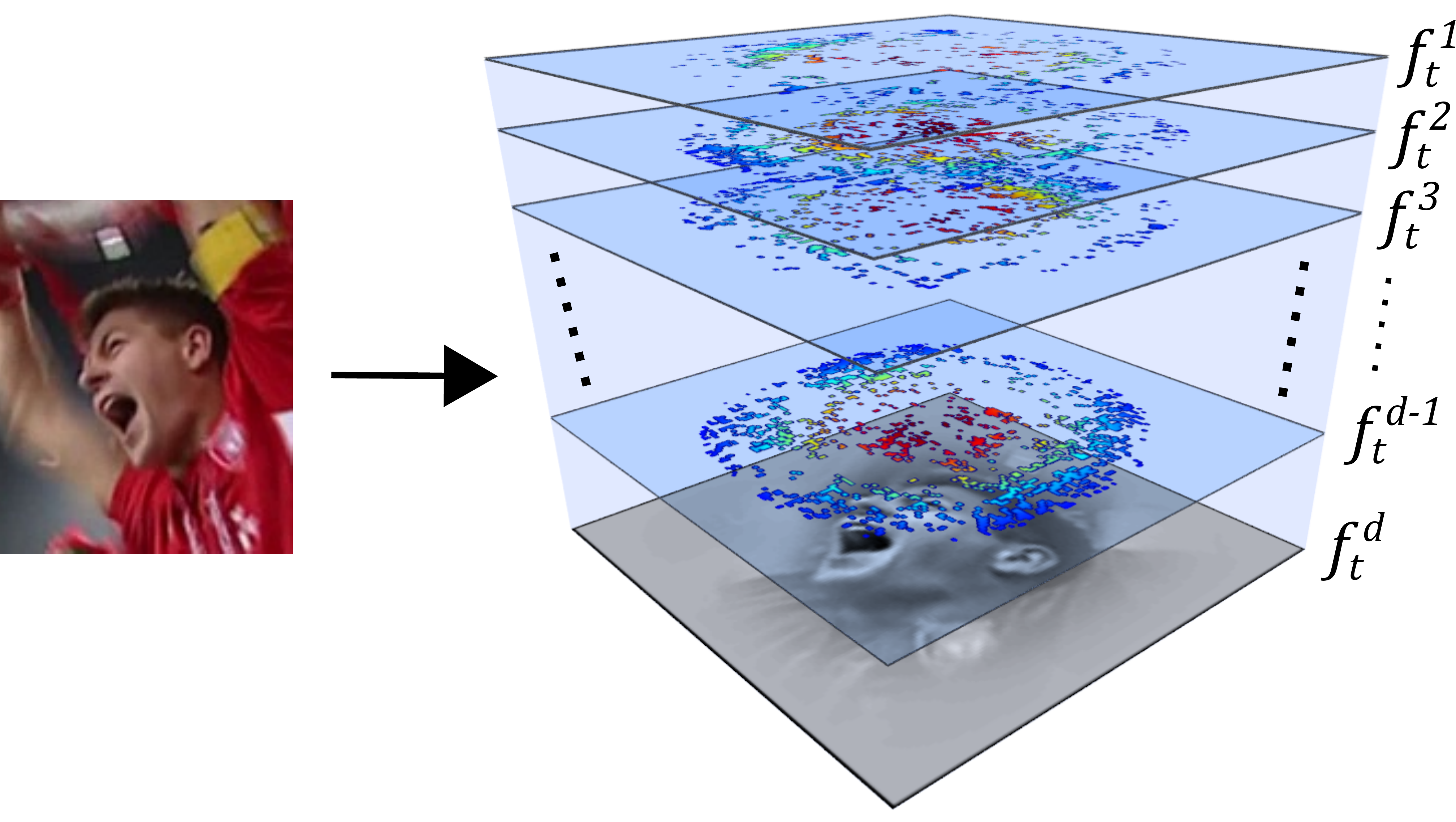}\label{fig:trans_sample}}\hspace{1.5cm}
   \subfloat[Scale filter sample.]{%
   \includegraphics[scale=0.4]{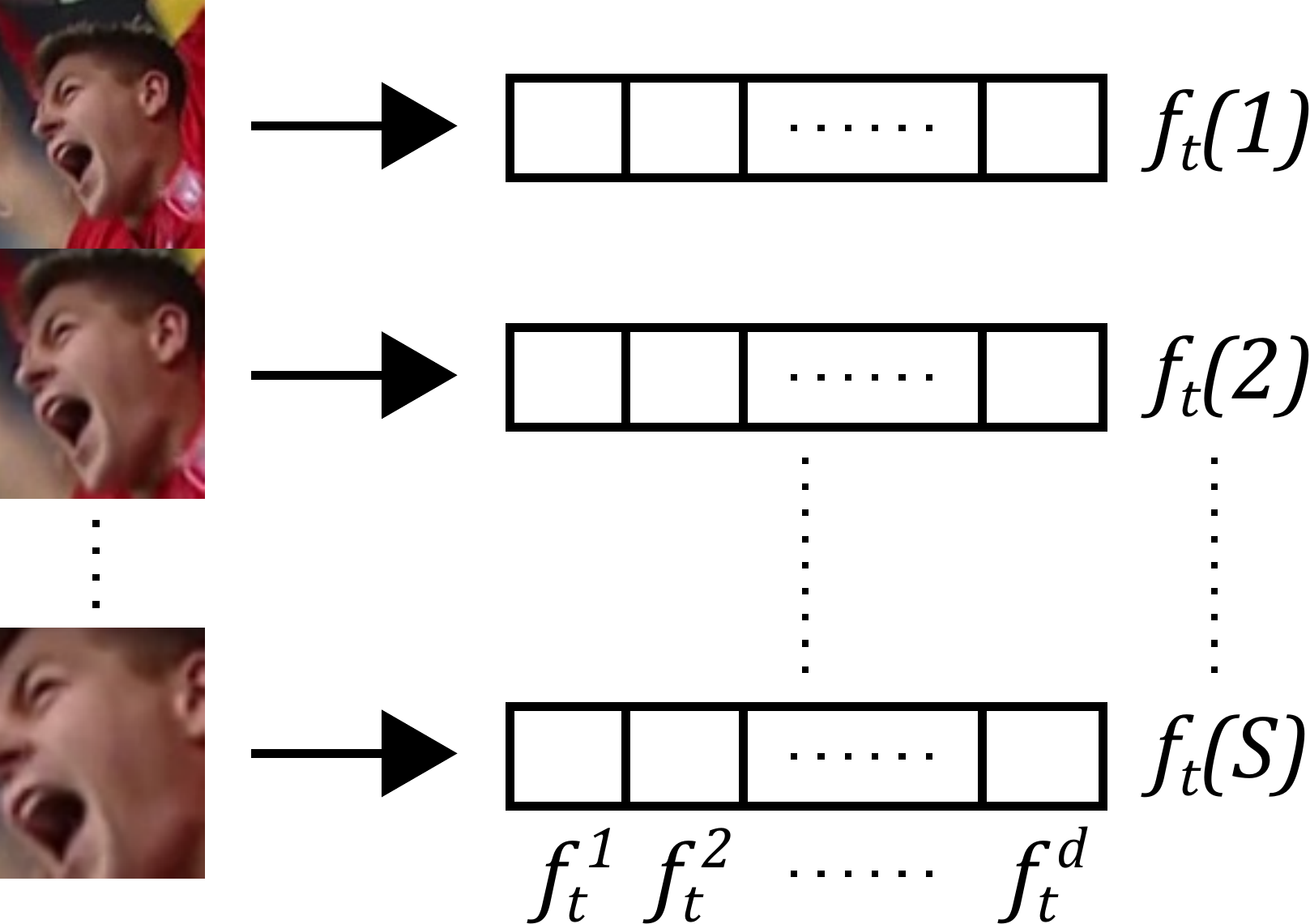}\label{fig:scale_sample}}
  \caption{Visualization of training samples used to update our DSST and fDSST approaches. The translation filter sample \protect\subref{fig:trans_sample} is extracted from a rectangular patch centered around the target. To update the scale filter \protect\subref{fig:scale_sample}, we first sample patches of the target appearance at a set of different scales. Each such patch is then mapped to a feature vector. The training sample used for updating the scale filter is set to the collection of these feature vectors.}
  \label{fig:training_samples}
\end{figure*}

As a baseline approach, we learn a 2-dimensional multi-channel DCF for translation-only tracking. The training and detection steps are performed as described in section~\ref{sec:MDCF}. Given the target location in frame number $t$, we first extract a training sample patch $f_t$ centered around the target (see figure~\ref{fig:trans_sample}). The translation filter is then updated using \eqref{eq:our_training}. To estimate the target location in a new frame $t$, a sample patch $z_t$ is first extracted at the previously estimated location. The correlation scores are then obtained by \eqref{eq:our_detection}.

\subsection{Multi-resolution Translation Filter}
\label{sec:multi_res_filter}
In object detection, a standard approach for detecting an object at different scales is to apply a classifier at multiple resolutions \cite{pedro10}. This strategy has been employed for the standard DCF-based tracker \cite{Li2014}. We learn a 2-dimensional translation filter using the same procedure as for the standard DCF tracker (see section~\ref{sec:baseline}). In the detection step, several patches at different resolutions are sampled centered around the previous target location. The translation filter is then applied to each patch independently using \eqref{eq:our_detection}. The translation and scale of the target is obtained by finding the resolution (scale) and location with the highest correlation score among all patches.

\subsection{Joint Scale Space Filter}
\label{sec:scale_space_approach}
A straightforward strategy for incorporating scale estimation is to construct a 3-dimensional scale space filter. This 3-dimensional filter jointly estimates the translation and scale of the target. It is achieved by computing the correlation scores in a box-shaped region of a scale pyramid representation. Both translation and scale estimates are then achieved by maximizing this score. 

To update the joint scale space filter, we first construct a feature pyramid in a rectangular area around the given target location. The feature pyramid is constructed such that the target size at the current scale corresponds to the spatial filter dimensions $M \times N$. The training sample $f_t$ is set to the rectangular cuboid of size $M \times N \times S$ centered around the target location and scale. Here, $S$ corresponds to the filter size in the scale dimension. The joint scale space filter is updated with \eqref{eq:our_training}, using a 3-dimensional Gaussian function as the desired correlation output $g$. The construction of the training sample is visualized in figure~\ref{fig:joint_sample}.

In the detection step, a feature pyramid is first constructed around the previously estimated target location and scale. The $M \times N \times S$ rectangular cuboid centered around this location is used as the test sample $z_t$. The correlation scores $y_t$ in this region of the scale space are then computed using \eqref{eq:our_detection}. 

\subsection{Iterative Joint Scale Space Filter}
\label{sec:iterative_filter}
As mentioned above, the feature pyramid at the detection step is constructed around the predicted target location. This might result in an inclusion of a shearing component in the transformation relating the test sample $z_t$ with the feature pyramid constructed around the actual target center. The shearing effect is caused by errors in the predicted target location.
This significantly affects the performance of the joint scale space filter by introducing a bias in the translation estimate.

The impact of the scale space shearing distortions can be reduced by iterating the detection step of the tracker. We therefore also evaluate an iterative joint scale space filter strategy. Given a new frame, we first apply the filter at the previous target location and scale. The target location is then updated with the location and scale corresponding to the maximum correlation score (as in section~\ref{sec:scale_space_approach}). The detection step is then iterated by constructing the feature pyramid around the current target estimate. This procedure is performed for a maximum number of iterations or until convergence is achieved. In most cases, the procedure converges since the shearing distortion of the pyramid is reduced as the location estimate improves. However, a major disadvantage of this approach is the added computational time due to the iterative detection procedure.

\section{Our Approach}
\label{sec:our_approach}
The scale extensions to the standard DCF tracker, described in section~\ref{sec:appraches}, significantly increase the computational cost of the tracker. Other than accuracy and robustness, the speed of the visual tracker is a crucial factor in many real world applications. Therefore, an ideal tracking approach should be accurate and robust while operating at real-time. Motivated by this observation, we propose a fast scale adaptive tracking approach. Our scale estimation approach is generic and can be used in any tracking framework with no scale estimation component.

\subsection{Discriminative Scale Space Tracking}
\label{sec:DSST}
We propose the Discriminative Scale Space Tracker (DSST), which is based on learning a separate 1-dimensional scale correlation filter. This scale filter can be applied at an image location to compute correlation scores in the scale dimension. These scores are then used to estimate the target scale. To construct the training sample $f_{t,\text{scale}}$, we extract features using variable patch sizes centered around the target. Let $P \times R$ denote the target size in the current frame and $S$ be the size of the scale filter. For each $n \in \left\{\left\lfloor-\frac{S-1}{2}\right\rfloor, \ldots, \left\lfloor\frac{S-1}{2}\right\rfloor\right\}$, we extract an image patch $I_n$ of size $a^nP \times a^nR$ centered around the target. Here, $a$ denotes the scale factor between feature layers. The value $f_{t,\text{scale}}(n)$ of the training sample $f_{t,\text{scale}}$ at scale level $n$, is set to the $d$-dimensional feature descriptor of $I_n$. The procedure for constructing the scale sample $f_{t,\text{scale}}$ is visualized in figure~\ref{fig:scale_sample}. Finally, \eqref{eq:our_training} is used to update the scale filter $h_{t,\text{scale}}$ with the new sample $f_{t,\text{scale}}$. In this case we use a 1-dimensional Gaussian as the desired correlation output $g$.

To estimate the translation of the target, we use the standard translation filter described in section~\ref{sec:baseline}. Typically, the target scale difference between two frames is small compared to the difference in translation. We therefore first apply the translation filter $h_{t,\text{trans}}$ given a new frame. The scale filter $h_{t,\text{scale}}$ is then applied at the new target location estimate. A scale estimation test sample $z_{t,\text{scale}}$ is extracted from this location using the same procedure as for the training sample $f_{t,\text{scale}}$. By maximizing the scale correlation scores \eqref{eq:our_detection}, we obtain the relative change in scale compared to the previous frame. Algorithm~\ref{alg:our_tracker} gives a brief outline of our DSST approach.

\begin{figure}[!t]
\algcaption{Our DSST approach: iteration at time step $t$.}
\label{alg:our_tracker}
\begin{boxedalgorithmic}
		\Require
			\Statex Image $I_t$.
			\Statex Previous target position $\bm{p}_{t-1}$ and scale $s_{t-1}$.
			\Statex Translation model $A_{t-1,\text{trans}}$, $B_{t-1,\text{trans}}$.
			\Statex Scale model $A_{t-1,\text{scale}}$, $B_{t-1,\text{scale}}$.
		\Ensure
			\Statex Estimated target position $\bm{p}_t$ and scale $s_t$.
			\Statex Updated translation model  $A_{t,\text{trans}}$, $B_{t,\text{trans}}$.
			\Statex Updated scale model $A_{t,\text{scale}}$, $B_{t,\text{scale}}$.
		\Statex
		\Statex \textbf{Translation estimation:}
		\State Extract sample $z_{t,\text{trans}}$ from $I_t$ at $\bm{p}_{t-1}$ and $s_{t-1}$.
		\State Compute correlation scores $y_{t,\text{trans}}$ using \eqref{eq:our_detection}.
		\State Set $\bm{p}_t$ to the target position that maximizes $y_{t,\text{trans}}$.
		\Statex
		\Statex \textbf{Scale estimation:}
		\State Extract sample $z_{t,\text{scale}}$ from $I_t$ at $\bm{p}_t$ and $s_{t-1}$.
		\State Compute correlation scores $y_{t,\text{scale}}$ using \eqref{eq:our_detection}.
		\State Set $s_t$ to the target scale that maximizes $y_{t,\text{scale}}$.
		\Statex
		\Statex \textbf{Model update:}
		\State Extract samples $f_{t,\text{trans}}$ and $f_{t,\text{scale}}$ from $I_t$ at $\bm{p}_t$ and $s_t$.
		\State Update the translation model $A_{t,\text{trans}}$, $B_{t,\text{trans}}$ using \eqref{eq:our_training}.
		\State Update the scale model $A_{t,\text{scale}}$, $B_{t,\text{scale}}$ using \eqref{eq:our_training}.
\end{boxedalgorithmic}
\end{figure}

\subsection{Fast Discriminative Scale Space Tracking}
\label{sec:fDSST}
Here we investigate strategies for reducing the computational cost of the proposed DSST method. Two approaches for reducing the computations required in the learning and detection steps of the multi-channel DCF described in section~\ref{sec:MDCF}, are presented. These approaches are: sub-grid interpolation of correlation scores and reduction of the feature dimensionality using Principal Component Analysis (PCA), also known as the discrete Karhunen-Lo\`eve transform.

\subsubsection{Sub-grid Interpolation of Correlation Scores}
Sub-grid interpolation allows us to use coarser feature grids for the training and detection samples. This affects the computational cost by reducing the size of the performed FFTs required to evaluate \eqref{eq:our_training} and \eqref{eq:our_detection} for training and detection respectively. We employ interpolation with trigonometric polynomials \cite{Oppenheim1996}. This is especially suitable since the  DFT coefficients of the correlation score, required to perform the interpolation, are already computed in \eqref{eq:our_detection}. The interpolated scores $\hat{y}_t$ are obtained by zero-padding the high frequencies of $Y_t$ in \eqref{eq:our_detection} such that its size is equal to the size of the interpolation grid. The interpolated scores $\hat{y}_t$ are then obtained by performing the inverse DFT of the padded $Y_t$.

\subsubsection{Dimensionality Reduction}
\label{sec:dim_reduction}
The computational cost of the DSST is dominated by the FFT. In our approach, the number of FFT computations scales linearly with the feature dimension $d$, since the training \eqref{eq:our_training} and detection \eqref{eq:our_detection} steps require one FFT per feature dimension. To reduce the required number of FFT computations, we employ a dimensionality reduction strategy. Similar to Danelljan et al.\ \cite{DanelljanCVPR14}, we base our dimensionality reduction scheme on the standard PCA. However, thanks to the simplicity of the linear kernel applied in this work, the smooth subspace update scheme \cite{DanelljanCVPR14} is not required.

To be able to reduce the number of FFT computations, we instead update a target template $u_t = (1-\eta) u_{t-1} + \eta f_t$. By the linearity of the Fourier transform, the numerator \eqref{eq:our_training_num} of the learned filter can then equivalently be obtained by $A_t^l = \overline{G} \ft \{ u_t^l \}$. 
The learned template $u_t$ is used to construct a projection matrix $P_t$. This matrix defines the low-dimensional subspace onto which the features are projected. The projection matrix $P_t$ is $\tilde{d} \times d$, where $\tilde{d}$ is the dimensionality of the compressed feature representation. We obtain $P_t$ by minimizing the reconstruction error of the target template $u_t$
\begin{equation}
	\label{eq:PCA_cost}
	\varepsilon = \sum_{\bm{n}} \left\| u_t(\bm{n}) - P_t \tp P_t u_t(\bm{n}) \right\|^2 .
\end{equation}
Here, the index tuple $\bm{n}$ ranges over all elements in the template $u_t$. Eq.~\eqref{eq:PCA_cost} is minimized under the orthonormality constraint $P_t P_t \tp = I$. A solution  is obtained by performing an eigenvalue decomposition of the auto-correlation matrix
\begin{equation}
	\label{eq:corr_mat}
	C_t = \sum_{\bm{n}} u_t(\bm{n}) u_t(\bm{n}) \tp .
\end{equation}
The rows of $P_t$ is set to the $\tilde{d}$ eigenvectors of $C_t$ corresponding to the largest eigenvalues.

The filter is updated using the compressed training sample $\tilde{F}_t = \ft \{P_t f_t\}$ and compressed target template $\tilde{U}_t = \ft \{P_t u_t\}$ as
\begin{subequations}
	\label{eq:our_PCA_training}
	\begin{align}
		\label{eq:our_PCA_training_num}
		\tilde{A}_t^l &= \overline{G} \tilde{U}_t^l \; , \quad l = 1, \ldots, \tilde{d} \\
		\label{eq:our_PCA_training_den}
		\tilde{B}_t &= (1 - \eta) \tilde{B}_{t-1} + \eta \sum_{k=1}^{\tilde{d}} \overline{\tilde{F}^k_t} \tilde{F}_t^k .
	\end{align}
\end{subequations}
The linear operation of $P_t$ is applied as an element-wise matrix multiplication $(P_t u_t)(\bm{n}) = P_t u_t(\bm{n})$ that projects the feature vector $u_t(\bm{n}) \in \reals^d$ onto the rows of $P_t$. As discussed above, the numerator \eqref{eq:our_PCA_training_num} can be obtained directly from the compressed template $\tilde{U}_t$, due to its linear relationship with the training samples $f_t$. However, the same strategy is not applicable for the denominator \eqref{eq:our_PCA_training_den}, since it depends on the auto-correlation of the training samples. Therefore, a different projection matrix $P_t$ is used for each term in $\tilde{B}_t$. Note that the aim of our dimensionality reduction is to approximate the denominator $B_t$ in the original update scheme \eqref{eq:our_training_den}. The training samples only appear as sums $\sum_{k=1}^{d} \overline{F^k_t} F_t^k$ in the denominator $B_t$. This corresponds to an inner product in the feature space, which is approximated by the inner product $\sum_{k=1}^{\tilde{d}} \overline{\tilde{F}^k_t} \tilde{F}_t^k$ evaluated in the subspace defined by $P_t$. Thus, a better approximation of $B_t$ is achieved by adapting the projection matrix for each frame $t$ and using different projections for each term $\sum_{k=1}^{\tilde{d}} \overline{\tilde{F}^k_t} \tilde{F}_t^k$.

The correlation scores at the test sample $z_t$ are obtained similarly to \eqref{eq:our_detection}, by applying the filter on the compressed sample $\tilde{Z}_t = \ft \{P_{t-1} z_t\}$,
\begin{equation}
	\label{eq:our_PCA_detection}
	Y_t = \frac{\sum_{l=1}^{\tilde{d}} \overline{\tilde{A}^l_{t-1}} \tilde{Z}^l_t}{\tilde{B}_{t-1} + \lambda} .
\end{equation}

\subsubsection{Compressed Scale Filter}
\label{sec:compressed_scale}
For the translation filter employed in our tracking approach, the feature dimensionality is smaller than the number of elements $u_{t,\text{trans}}(\bm{n})$ in the template. For the scale filter however, the opposite is true. In this case the rank of the auto-correlation matrix \eqref{eq:corr_mat} is lesser than or equal to the number of scales, i.e.\ $\text{rank}(C_{t,\text{scale}}) \leq S$. This is evident, since the scale template $u_{t,\text{scale}}$ consists of a $d$-dimensional feature vector $u_{t,\text{scale}}(n) \in \reals^d$ for each scale index $n \in \left\{\left\lfloor-\frac{S-1}{2}\right\rfloor, \ldots, \left\lfloor\frac{S-1}{2}\right\rfloor\right\}$. In the scale filter case, the feature dimensionality $d \approx 1000$ is far larger than the number of PCA samples $S = 17$ (see section~\ref{sec:parameters} for details concerning features and parameters). The scale template $u_{t,\text{scale}}$ can thus be compressed to $\tilde{d} = S$ feature dimensions without any loss of information. The same holds true for the scale filter training sample $f_{t,\text{scale}}$.

The presented dimensionality reduction technique is adapted to the scale filter by utilizing the aforementioned properties. For efficiency, two projection matrices $P_{t,\text{scale}}^u$ and $P_{t,\text{scale}}^f$ are computed based on $u_{t,\text{scale}}$ and $f_{t,\text{scale}}$ respectively. The template and sample can then be compressed without loss of information using $\tilde{u}_{t,\text{scale}} = P_{t,\text{scale}}^u u_{t,\text{scale}}$ and $\tilde{f}_{t,\text{scale}} = P_{t,\text{scale}}^f f_{t,\text{scale}}$. These compressed versions are then used to update the scale filter as in \eqref{eq:our_PCA_training}. The scale filter \eqref{eq:our_PCA_training} is not affected by the dimensionality reduction in this case, since both the template and the training sample can be exactly reconstructed as $u_{t,\text{scale}} = (P_{t,\text{scale}}^u)\tp \tilde{u}_{t,\text{scale}}$ and $f_{t,\text{scale}} = (P_{t,\text{scale}}^f)\tp \tilde{f}_{t,\text{scale}}$ respectively. The same holds true for the Fourier coefficients $U_{t,\text{scale}}$ and $F_{t,\text{scale}}$ by linearity. To compute the scale correlations scores at the detection stage, we apply \eqref{eq:our_PCA_detection} using a compressed test sample $\tilde{z}_{t,\text{scale}} = P_{t-1,\text{scale}}^u z_{t,\text{scale}}$.

For computational and memory efficiency purposes, we do not explicitly construct the auto-correlation matrix \eqref{eq:corr_mat}, but rather obtain the projection matrices $P_{t,\text{scale}}^u$ and $P_{t,\text{scale}}^f$ through a QR-factorization of $u_{t,\text{scale}}$ and $f_{t,\text{scale}}$ respectively. This does not affect the tracking output since the multi-channel DCF approach (as presented in section~\ref{sec:MDCF}) is invariant to any change of orthonormal basis in the feature representation.

\subsubsection{Search Space Expansion for Robustness}
The strategies presented above are employed in both the translation and scale filter of our tracker, to give a significant reduction in computational cost. This enhancement provides the flexibility to use a larger target translation search space. The search space expansion is performed by increasing the size of the translation filter. Note that such an expansion significantly increases the computational time of other DCF based trackers the increased size of the performed FFTs. By employing the strategies proposed strategies above, the size of the FFTs and number of FFTs are reduced sufficiently in order to increase the filter size without sacrificing real-time performance. 

The expansion in the filter size provides increased contextual information in the filter. It also helps in alleviating the problem of fast motion and occlusions, by using a larger search space. In summary, by incorporating the strategies proposed in this section, our fast DSST (fDSST) method improves the robustness of the tracker, while operating at twice the speed of the DSST.

\section{Experiments}
\label{sec:experiments}
We perform extensive evaluations on two benchmark datasets to validate our approach. Section~\ref{sec:parameters} presents the details about features and parameters used in our experiments. In section~\ref{sec:experimental_setup}, we describe the used benchmark datasets and evaluation protocols. The DCF based scale estimation approaches, discussed earlier, are compared in section~\ref{sec:baseline_experiment}. In section~\ref{sec:fDSST_comp} the extended fast version (fDSST) of our approach is compared to the standard DSST. A comparison with state-of-the-art methods on the OTB dataset is given in section~\ref{sec:sota_comp}. Finally, we present results on the VOT 2014 dataset in section~\ref{sec:vot2014}.

\subsection{Implementation Details}
\label{sec:parameters}
We set the regularization parameter to $\lambda = 0.01$ and the learning rate to $\eta = 0.025$. The standard deviation of the desired correlation output $g$ is set to $1/16$ of the target size in the translation dimensions. Note that this set of parameters is used for all DCF-based trackers presented in section~\ref{sec:appraches} and \ref{sec:our_approach} to achieve a fair comparison. For all presented trackers except our fDSST, we also use the same spatial size $M \times N$ of the filter, which is set to twice the initial target size. Thanks to the strategies presented in section~\ref{sec:fDSST} for improving the tracking speed, we can use larger filter for our fDSST without sacrificing real-time performance. For the fDSST we therefore set the size of the translation filter to three times the initial target size. 

For the joint scale space filters (sections~\ref{sec:scale_space_approach} and \ref{sec:iterative_filter}) and our DSST (section~\ref{sec:DSST}) we use $S = 33$ number of scales. For the fDSST (section~\ref{sec:fDSST}) we interpolate the scale correlation output from $S = 17$ to $\hat{S} = 33$ scales using the described approach. The joint and the discriminative approaches all use a scale factor of $a = 1.02$, and the standard deviation in the scale dimension of the desired correlation output $g$ is set to $1/16$ times the number of scales $S$. For the multi-resolution translation filter (section~\ref{sec:multi_res_filter}), we observed that setting the number of scales $S$ or the scale factor $a$ similarly to the other scale adaptive approaches (i.e.\ $S=33$ and/or $a = 1.02$) gave inferior performance in our experiments. For this approach we therefore use $S = 5$ scales and a scale factor of $a = 1.005$, which turned out to be the best setting in our experiments.

We employ PCA-HOG \cite{pedro10} for image representation, with the implementation provided by \cite{PMT}. The same pixel-dense feature representation is used for all trackers presented in section~\ref{sec:appraches} and for the translation filter in our DSST. This feature is obtained by augmenting HOG computed with $1 \times 1$ pixel cells with the image intensity (grayscale) value. To save computations, we use a coarser feature grid for the translation filter in our fDSST. To achieve pixel-dense correlation scores we then apply the interpolation technique described in section~\ref{sec:fDSST}. The feature vector is constructed using HOG with $4 \times 4$ cells. This HOG vector is augmented with the average grayscale value in the corresponding cell. The grayscale features are always normalized to the range $[-\frac{1}{2}, \frac{1}{2}]$.

For the scale filter in our DSST and fDSST, we compute the feature descriptor of the image patch $I_n$ by first re-sizing the patch to a fixed size. HOG features are then extracted using a cell size of $4 \times 4$. The fixed patch size is set to the initial target size. However, for targets with an initial area larger than $512$ pixels, we calculate a fixed size with a preserved aspect ratio and an area of $512$ pixels. This ensures a maximum feature descriptor length of $992$. For all filters (translation, scale and joint), each feature channel in the extracted sample is always multiplied by a Hann window, as described in \cite{MOSSE2010}.

The fDSST applies PCA as described in section~\ref{sec:dim_reduction} to reduce the dimensionality of the translation filter. The 32-dimensional HOG and intensity combination is reduced to 18 dimensions in our experiments. For the scale filter, we apply the modified reduction scheme presented in section~\ref{sec:compressed_scale}. This approach reduces the dimensionality of the scale features from $d \approx 1000$ to only $S = 17$ dimensions.

\begin{table}[!t]
	\caption{A comparison of our DSST approach with the baseline translation filter and the DCF based scale adaptive trackers discussed in section~\ref{sec:appraches}. The mean overlap precision (OP) ($\%$) and distance precision (DP) ($\%$) over all the 50 videos in the OTB dataset are presented. We also report the tracker speed in mean FPS. The two best results are shown in red and blue fonts respectively. Our DSST approach significantly outperforms all compared trackers, while being the fastest among the scale adaptive methods.}
	\centering
	\begin{tabular}{lccc}
\toprule
&Mean OP&Mean DP&Mean FPS\\\midrule
Translation DCF&57.7&70.8&\textbf{\textcolor{red}{57.3}}\\
Multi-Resolution DCF&\textit{\textcolor{blue}{65.2}}&\textit{\textcolor{blue}{74.8}}&16.9\\
Joint DCF&63.2&72.1&1.46\\
Iterative Joint DCF&64.1&74.2&1.01\\
\textbf{DSST} (ours)&\textbf{\textcolor{red}{67.7}}&\textbf{\textcolor{red}{75.7}}&\textit{\textcolor{blue}{25.4}}\\\bottomrule
\end{tabular}

	\label{tab:baseline_mean_summary}
\end{table}

\subsection{Experimental Setup}
\label{sec:experimental_setup}
Our approaches are implemented in Matlab. All experiments are performed on an Intel Xeon 2 core 2.66 GHz CPU with 16 GB RAM. For the tracking approaches presented in section~\ref{sec:appraches} and \ref{sec:our_approach}, the same parameter settings are used for all experiments and videos. Our methods are quantitatively evaluated on the Online Tracking Benchmark (OTB) dataset, following the evaluation protocol described in \cite{Wu13g}. This dataset contains 50 challenging image sequences. We also evaluate our method on the VOT 2014 dataset \cite{VOT2014}.

The tracking results on the OTB dataset are reported using three standard evaluation metrics, namely overlap precision (OP), distance precision (DP) and tracking speed in frames per second (FPS). The OP score is computed as the percentage of frames in a video where the intersection-over-union overlap with the ground truth exceeds a certain threshold. In the tables we report the OP at a threshold of $0.5$, which corresponds to the PASCAL evaluation criterion. The DP score is defined as the percentage of frames in a video where the Euclidean distance between the tracking output and ground truth centroids is smaller than a threshold. A threshold of $20$ pixels is used in this work \cite{Wu13g,Henriques12d}.

We also provide \emph{success plots} of the results on the OTB dataset. In the success plot, the mean OP over all videos is plotted against the range of overlap thresholds $[0,1]$. In the legend we report the \emph{area-under-the-curve} (AUC) score for each tracker.

\begin{figure}[!t]
	\centering
	\newcommand{\wid}{0.4\textwidth}%
	\includegraphics[width=\wid]{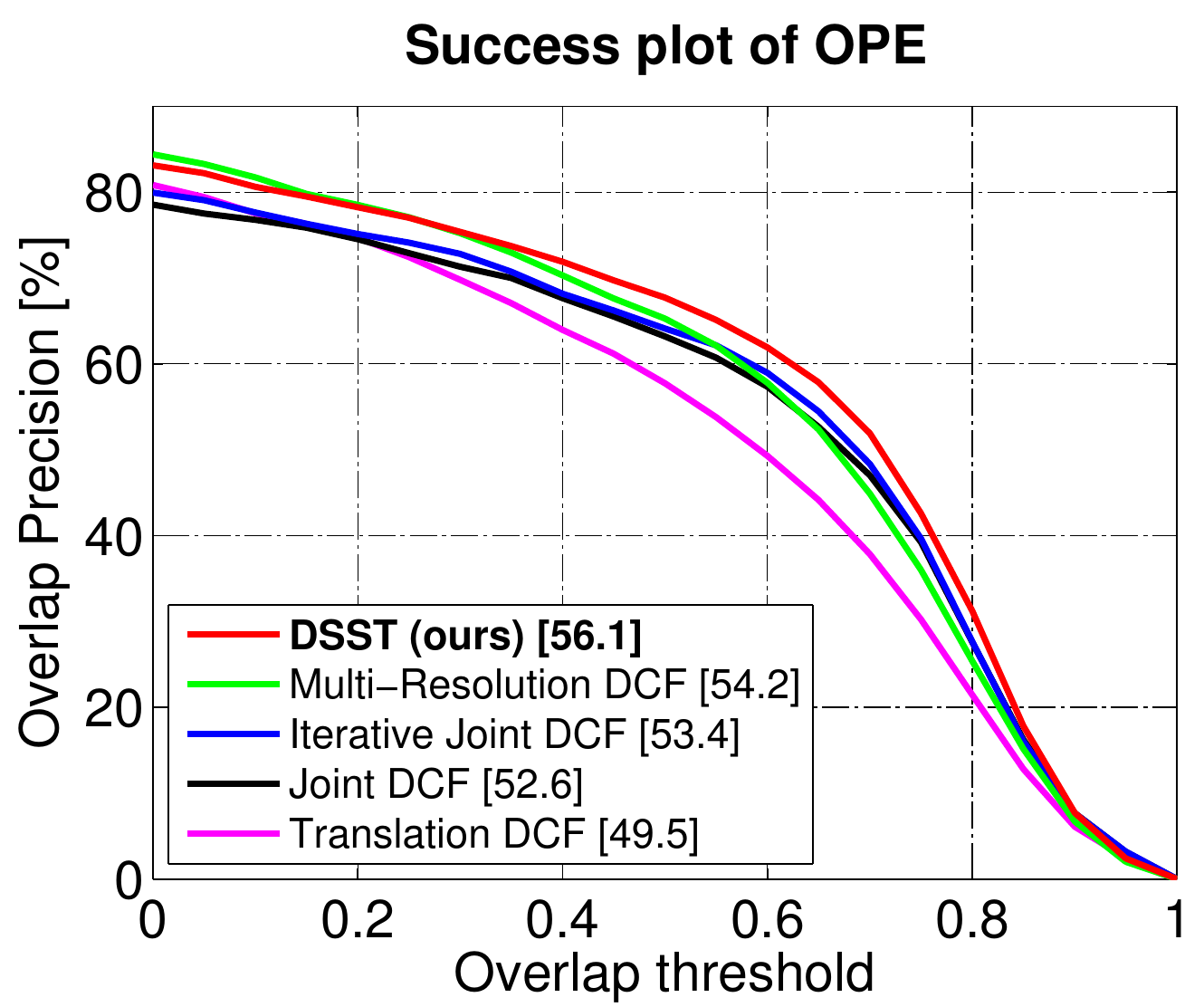}
	\caption{Success plot showing a comparison of the different DCF-based tracking approaches over all the 50 videos in the OTB dataset. The legend of the success plot contains the area-under-the-curve (AUC) score for each method. The best results are obtained with our DSST method, improving the baseline translation DCF tracker by $6.6 \%$ in AUC.}
	\label{fig:baseline_success}
\end{figure}

\subsection{Experiment 1: DCF-based Scale Estimation}
\label{sec:baseline_experiment}
Table~\ref{tab:baseline_mean_summary} shows a comparison of the DCF-based tracking approaches presented in section~\ref{sec:appraches} and \ref{sec:our_approach}, on the OTB dataset. The standard translation based DCF tracker obtains a mean OP of $57.7 \%$. The joint scale space filter and its iterative extension achieve a mean OP of $63.2 \%$ and $64.1 \%$ respectively. The best results are obtained using our DSST method, providing a significant gain of $10.0 \%$ compared to the baseline translation tracker. Similarly, our approach also provides improved performance in mean DP compared to the other scale adaptive trackers. It worth mentioning that our DSST provides significantly better performance while being 17 times faster compared to the joint scale space filter tracker. Similarly, the DSST method achieves higher accuracy compared to the multi-resolution translation filter while operating at a higher frame-rate. 

Figure~\ref{fig:baseline_success} show the success plot illustrating the mean overlap precision over all the 50 videos in the OTB dataset. The standard translation filter achieves an AUC score of $49.5 \%$. The joint scale space filter improves the performance with an AUC score of $52.6 \%$. Our DSST approach further improves the performance by providing a significant gain of $6.6 \%$ compared to the baseline translation tracker.

We also perform a baseline comparison of the DCF trackers presented in section~\ref{sec:appraches} and \ref{sec:our_approach},  on the VOT 2014 dataset. In VOT 2014, the methods are compared both in terms of accuracy and robustness. The accuracy is calculated as the number of frames where the overlap is above a certain threshold. The robustness score is based on the number of times the tracker fails in a video. A tracker is always restarted five frames after a tracking failure occurs. Contrary to the OTB dataset, the ground truth bounding boxes in the VOT 2014 dataset are not axis aligned. The final tracker scores are based on how the trackers are ranked in terms of accuracy and robustness in each video. We refer to \cite{VOT2014} for more details about the VOT evaluation protocol.

\begin{table}[!t]
	\caption{A baseline comparison on VOT 2014 of our DSST approach with the translation-only DCF and the scale adaptive DCF trackers discussed in section~\ref{sec:appraches}. We report accuracy and robustness ranks, along with the final averaged ranking score. The average overlap and failures over the videos are also shown in the last two columns. Our DSST obtains the best final rank and outperforms the compared trackers in terms of robustness with an average failure rate of $1.16$.}
	\centering
	\resizebox{\columnwidth}{!}{\begin{tabular}{l@{~~}c@{~~~}c@{~~~}c@{~~~}c@{~~~}c@{~~~}}
\toprule
&Acc. Rank&Rob. Rank&Final Rank&Overlap&Failures\\\midrule
\textbf{DSST} (ours)&2.88&2.72&\textbf{\textcolor{red}{2.80}}&0.62&1.16\\
Iterative Joint DCF&2.80&3.04&\textit{\textcolor{blue}{2.92}}&0.63&1.56\\
Joint DCF&2.80&3.14&2.97&0.62&1.60\\
Multi-Resolution DCF&3.00&2.96&2.98&0.62&1.40\\
Translation DCF&3.49&3.14&3.31&0.55&1.48\\\bottomrule
\end{tabular}
}
	\label{tab:baseline_vot}
\end{table}

Table~\ref{tab:baseline_vot} shows the ranking scores on VOT 2014, along with the average overlap and failures. The translation alone tracker achieves a final rank of $3.31$. Among scale adaptive DCF approaches, our DSST provides the best results with a final rank of $2.80$. Further, our approach outperforms the compared trackers by significantly reducing the average failure rate to $1.16$, while maintaining the accuracy.

In summary, our approach significantly improves the performance of the standard translation tracker. This shows that accurate scale estimation is crucial for the robustness of the tracker. Our approach also provides superior performance and frame-rate compared to the other scale adaptive DCF-based trackers.

\begin{table}[!b]
	\centering
  \caption{A comparison of our discriminative scale space trackers (DSST and fDSST). The mean overlap precision (OP) ($\%$) and distance precision (DP) ($\%$) over all the 50 videos in the OTB dataset are presented. The best results are displayed in red. Our fDSST achieves significantly better results while operating at double mean FPS.}
    \begin{tabular}{lccc}
\toprule
&Mean OP&Mean DP&Mean FPS\\\midrule
DSST&67.7&75.7&25.4\\
fDSST&\textbf{\textcolor{red}{74.3}}&\textbf{\textcolor{red}{80.2}}&\textbf{\textcolor{red}{54.3}}\\\bottomrule
\end{tabular}

  \label{tab:fdsst_mean_summary}
\end{table}

\subsection{Experiment 2: Fast Discriminative Scale Space Tracker}
\label{sec:fDSST_comp}
Here we compare our standard discriminative scale space tracker (DSST) presented in section~\ref{sec:DSST} with the extended fast version (fDSST) presented in section~\ref{sec:fDSST}. In addition to the proposed strategies for increasing the tracker speed, the fDSST also employs a larger target search space to improve robustness. Table~\ref{tab:fdsst_mean_summary} shows a comparison of the two approaches on the OTB dataset. Our fDSST approach improves the performance by providing a gain of $7.0 \%$ and $4.4 \%$ in mean OP and DP respectively. Furthermore, this significant gain in performance is achieved while operating at over twice the speed in mean FPS compared to the DSST approach.

\begin{figure}[!t]
	\centering
	\newcommand{\wid}{0.35\textwidth}
	\includegraphics[width=\wid]{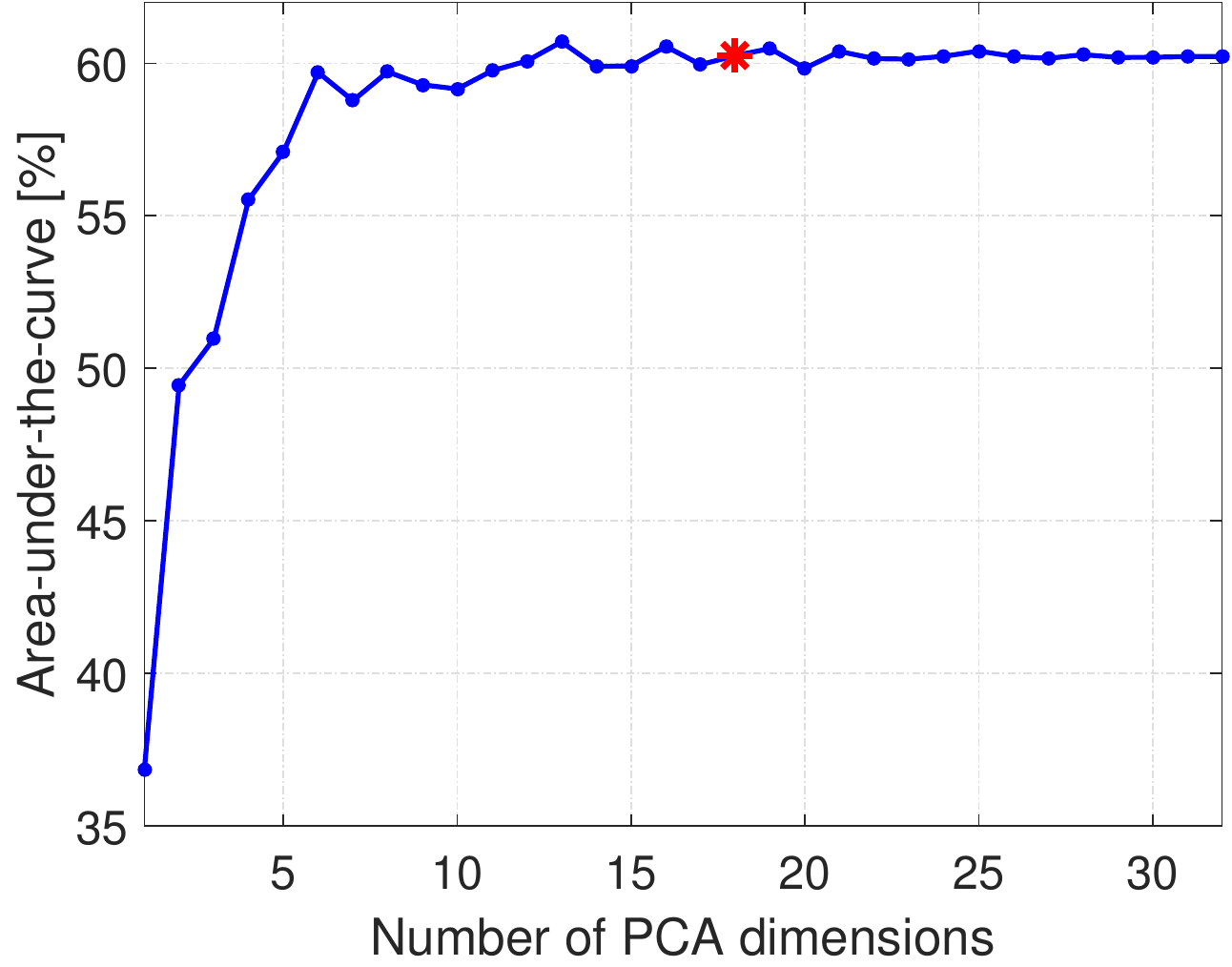}
	\caption{Impact of the number of subspace dimensions for the translation filter in the fDSST. Tracking performance on the OTB dataset in terms of area-under-the-curve (AUC) score is plotted for different choices of the dimensionality $\tilde{d}$. We use 18 dimensions (red) in our experiments.}
	\label{fig:pca}
\end{figure}

As described in section~\ref{sec:dim_reduction}, we employ a dimensionality reduction scheme in our fDSST framework. We analyze the impact of varying the number of subspace dimensions for the translation filter, on the OTB dataset. Figure~\ref{fig:pca} shows the tracking performance, in AUC, for different choices of this parameter. The performance of our fDSST largely remains consistent when the number of dimensions is reduced from 32 and then degrades rapidly at about 6. Our results suggest that the feature dimensionality can be significantly reduced with our framework, while preserving tracking performance. To achieve consistent and stable results, we set the number of PCA dimensions to 18 for all our experiments.

\begin{table*}[!t]
	\caption{A comparison of our fDSST approach with 18 state-of-the-art trackers. The mean overlap precision (OP) ($\%$) and distance precision (DP) ($\%$) over all the 50 videos in the OTB dataset are presented. The two best results are displayed in red and blue fonts respectively. Our approach achieves superior performance compared to the existing trackers.}
	\centering
	\resizebox{\textwidth}{!}{\begin{tabular}{l@{~~~}c@{~~~}c@{~~~}c@{~~~}c@{~~~}c@{~~~}c@{~~~}c@{~~~}c@{~~~}c@{~~~}c@{~~~}c@{~~~}c@{~~~}c@{~~~}c@{~~~}c@{~~~}c@{~~~}c@{~~~}c@{~~~}c@{~~~}c@{~~~}c@{~~~}}
\toprule
&IVT&MIL&CT&TLD&DFT&EDFT&L1APG&CSK&LOT&CPF&CXT&Frag&LSST&LSHT&ASLA&SCM&KCF&Struck&SAMF HOG&SAMF&\textbf{fDSST}\\\midrule
Mean OP&42.7&36.5&24.8&48.9&44.4&49.8&44.0&44.2&43&38.9&47.7&39.7&41.8&47.0&56.4&53.0&62.3&58.8&66.6&\textit{\textcolor{blue}{69.7}}&\textbf{\textcolor{red}{74.3}}\\
Mean DP&50.6&45.5&32.4&55.3&49.6&56.7&48.5&54.4&52.7&48.8&55.5&46.1&50.2&56.9&59.2&56.3&74.0&68.7&73.7&\textit{\textcolor{blue}{77.7}}&\textbf{\textcolor{red}{80.2}}\\
Mean FPS&12.3&12.5&67.5&23.6&9.83&21.9&1.12&\textbf{\textcolor{red}{198}}&0.50&68.2&11.3&4.40&3.69&12.4&1.04&0.09&\textit{\textcolor{blue}{174}}&10.4&15.5&14.9&54.3\\\bottomrule
\end{tabular}
}
	\label{tab:sota_mean_summary}
\end{table*}
\begin{figure*}[!t]
	\centering
	\newcommand{\wid}{0.32\textwidth}
	\subfloat[One pass evaluation (OPE).]{%
		\includegraphics[width=\wid]{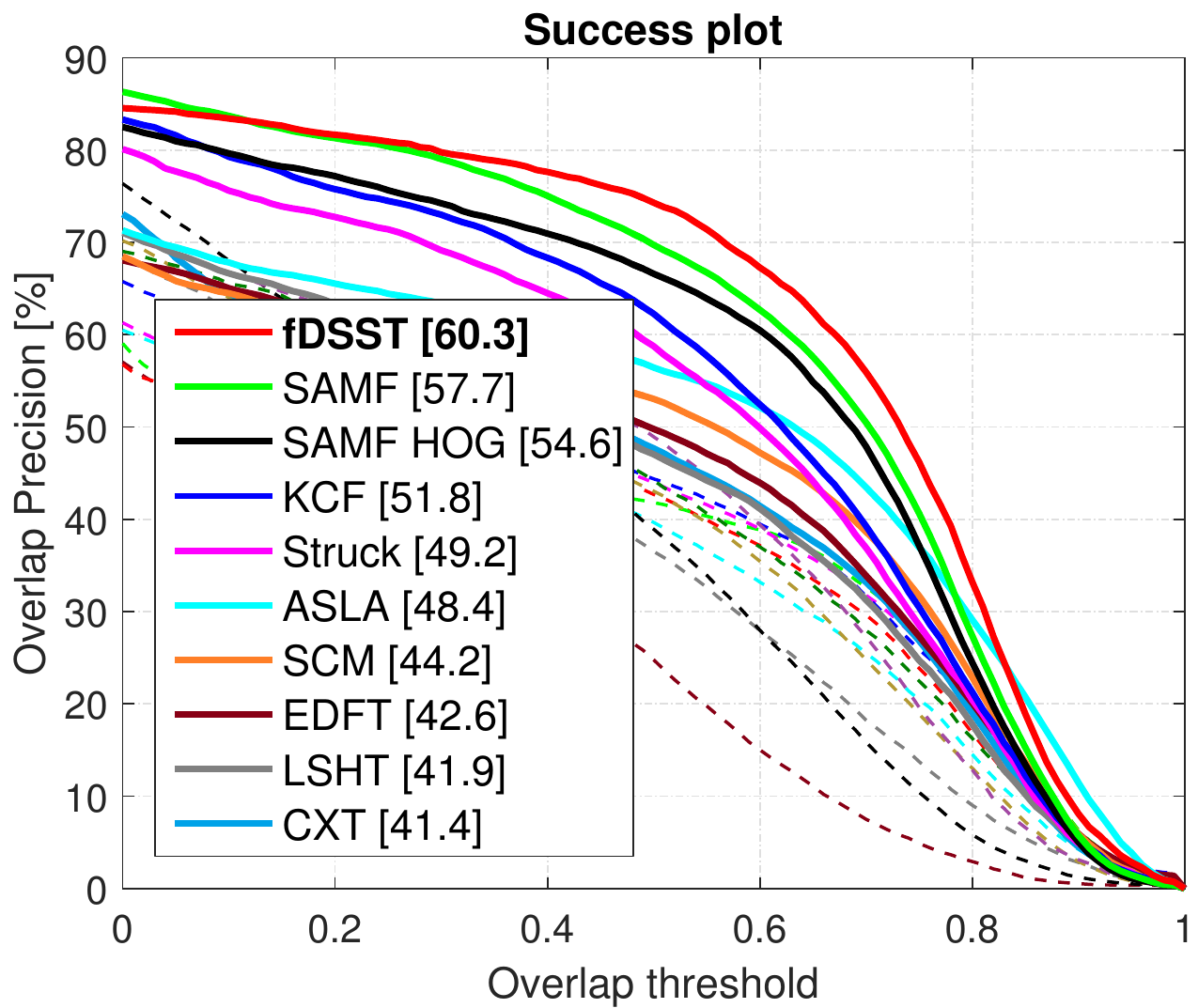}%
		\label{fig:sota_ope}}\hspace{1mm}
	\subfloat[Temporal robustness evaluation (SRE).]{%
		\includegraphics[width=\wid]{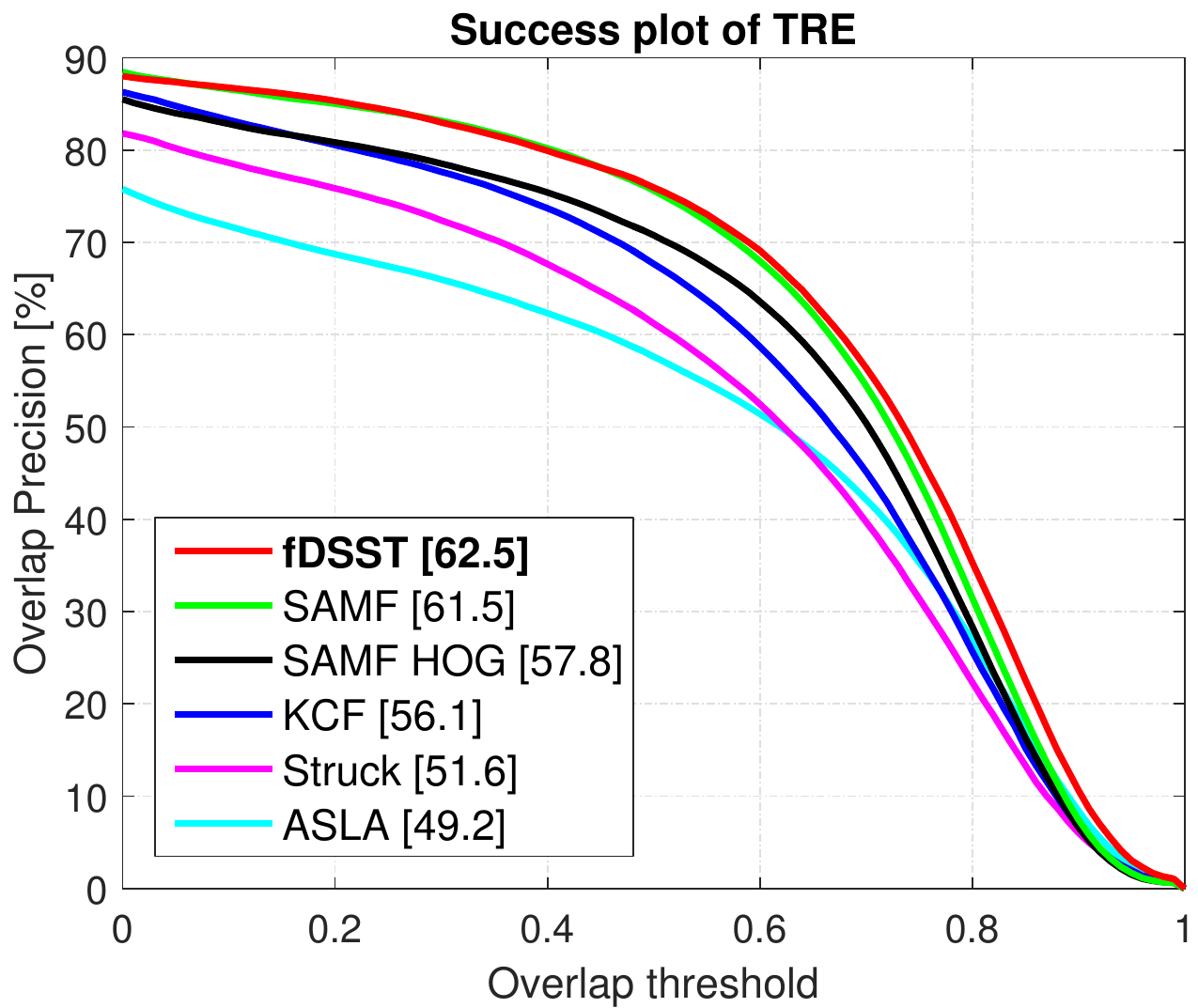}%
		\label{fig:TRE}}\hspace{1mm}
	\subfloat[Spatial robustness evaluation (SRE).]{%
		\includegraphics[width=\wid]{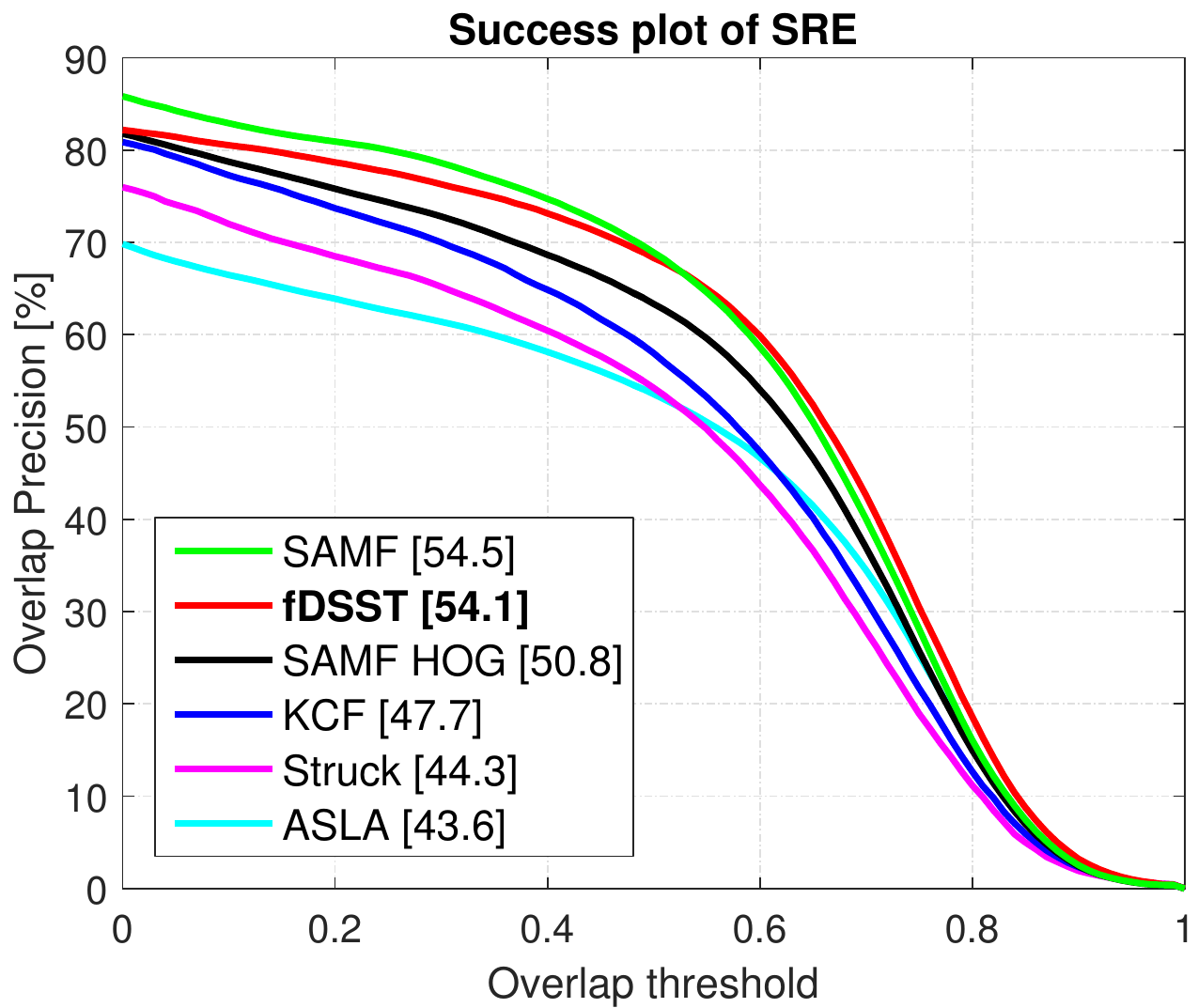}%
		\label{fig:SRE}}
	\caption{Success plot showing the performance of our fDSST compared to several state-of-the-art approaches on the OTB dataset. The area-under-the-curve (AUC) score for each tracker is reported in the legend. For the OPE \protect\subref{fig:sota_ope}, only the top ten trackers are displayed in the legend for clarity. Our method outperforms the second best tracker (SAMF) with $2.6 \%$ in AUC. We also compare the temporal \protect\subref{fig:TRE} and spatial \protect\subref{fig:SRE} robustness of our approach with the top 5 trackers. In both cases, our approach provides promising performance compared to existing tracking approaches.}
	\label{fig:OTB_SOTA}
\end{figure*}

\subsection{Experiment 3: State-of-the-Art Comparison}
\label{sec:sota_comp}
In this section, we provide a comprehensive comparison of our fDSST with 19 state-of-the-art methods in the literature. The trackers used for our comparison are: MIL \cite{Babenko09b}, IVT \cite{Ross08d}, CT \cite{Zhang12c}, TLD \cite{Mikolajczyk10d}, DFT \cite{Laura12d}, EDFT \cite{Felsberg13c}, ASLA \cite{Jia12d}, L1APG \cite{Bao12d}, CSK \cite{Henriques12d}, SCM \cite{Zhong12g}, LOT \cite{Oron12b}, CPF \cite{Perez02b}, CXT \cite{Dinh11}, Frag \cite{Adam6c}, Struck \cite{Torr11b}, LSHT \cite{Shengfeng13b}, LSST \cite{Wang13d}, KCF \cite{henriquesARXIV14} and SAMF \cite{Li2014}. For a fair comparison, we also compare with a version of SAMF (termed \emph{SAMF HOG}) that employs the same feature set as fDSST. For all methods except LSST, LSHT, EDFT, KCF and SAMF the code or binaries are provided with the OTB dataset \cite{Wu13g}.

Table \ref{tab:sota_mean_summary} provides a comparison using mean OP and DP over all 50 videos in the OTB dataset. We also provide a comparison of the speed of the trackers in mean FPS. The best two results are reported in red and blue fonts respectively. The scale adaptive SCM tracker, which is based on sparse representations, provides a mean OP of $53.0 \%$. The structural SVM based Struck tracker achieves a mean OP of $58.8 \%$. The KCF tracker \cite{henriquesARXIV14}, which is based on a kernelized correlation filter for translation estimation, obtains a mean OP of $62.3 \%$. The SAMF tracker extends the KCF with multiple features and scale estimation using a multi-resolution translation filter approach. It obtains the best results among the existing methods, with a mean OP of $69.7 \%$. Contrary to the SAMF tracker, our fDSST approach only employs intensity information, while using an explicit filter for scale estimation. Our tracker outperforms SAMF by $4.6 \%$ in mean OP. Similarly, our approach also provides superior results in terms of mean DP compared to the existing trackers. Finally, it is worth mentioning that our approach achieves superior performance while operating at real-time ($54.3$ in mean FPS). 

The success plot in figure~\ref{fig:sota_ope} shows the overlap precision (OP) over a range of overlap thresholds. The OP is calculated as the mean over all the 50 videos in the OTB dataset. For clarity, we only show the top 10 trackers in this comparison. Our approach significantly outperforms existing trackers, by achieving an AUC score of $60.3 \%$. It is worth mentioning that our approach provides a gain of $8.5 \%$ and $2.6 \%$ in AUC compared to the KCF and SAMF trackers respectively.

\begin{table*}[!t]
	\caption{Attribute-based comparison with state-of-the-art trackers on the OTB dataset. We report the AUC scores ($\%$) for the top ten trackers. The number of videos associated with the attribute is shown in parenthesis. Our approach provides improved performance on 7 out of 11 attributes. In scale variation videos, our method significantly outperforms all trackers including SAMF, which employs a multi-resolution scale estimation strategy.}
	\centering
	\resizebox{\textwidth}{!}{\begin{tabular}{l@{~}c@{~~~}c@{~~~}c@{~~~}c@{~~~}c@{~~~}c@{~~~}c@{~~~}c@{~~~}c@{~~~}c@{~~~}c@{~~~}}
\toprule
&Scale &Illumination &Out-of-plane&Occlusion&Background&Deformation&Motion&Fast&In-plane&Out of&Low\\
&variation (28)&variation (25)&rotation (39)&(29)&clutter (21)&(19)&blur (12)&motion (17)&rotation (31)&view (6)&resolution (4)\\\midrule
\textbf{fDSST}&\textbf{\textcolor{red}{57.1}}&\textbf{\textcolor{red}{59.8}}&\textbf{\textcolor{red}{57.2}}&\textit{\textcolor{blue}{56.0}}&\textbf{\textcolor{red}{62.4}}&56.9&\textbf{\textcolor{red}{59.8}}&\textbf{\textcolor{red}{55.9}}&\textbf{\textcolor{red}{58.3}}&\textit{\textcolor{blue}{56.3}}&\textit{\textcolor{blue}{39.9}}\\
SAMF&\textit{\textcolor{blue}{52.0}}&\textit{\textcolor{blue}{53.9}}&\textit{\textcolor{blue}{56.0}}&\textbf{\textcolor{red}{62.8}}&53.1&\textbf{\textcolor{red}{63.0}}&\textit{\textcolor{blue}{52.4}}&\textit{\textcolor{blue}{52.0}}&51.3&\textbf{\textcolor{red}{61.9}}&36.5\\
SAMF HOG&46.9&51.2&51.6&54.4&52.5&\textit{\textcolor{blue}{57.5}}&50.5&46.3&\textit{\textcolor{blue}{54.3}}&50.8&\textbf{\textcolor{red}{41.1}}\\
KCF&42.8&49.7&49.9&51.7&\textit{\textcolor{blue}{54.0}}&53.9&50.0&46.3&50.1&55.6&31.3\\
Struck&43.1&44.8&45.3&44.9&45.0&45.0&47.7&49.4&45.6&44.9&36.6\\
ASLA&49.7&49.8&46.9&44.7&50.5&47.6&31.9&29.8&45.2&40.8&16.8\\
SCM&48.1&40.3&42.4&42.1&42.9&37.0&26.8&30.3&41.3&35.6&30.2\\
EDFT&35.2&36.1&40.2&35.9&44.5&41.4&40.5&36.5&40.2&27.1&28.7\\
LSHT&35.5&40.1&42.1&38.9&41.4&41.0&27.8&29.2&40.5&38.0&11.8\\
CXT&38.5&34.6&40.3&36.2&31.8&31.7&34.2&36.3&42.5&40.2&27.7\\\bottomrule
\end{tabular}
}
	\label{tab:attribute}
\end{table*}

\begin{figure*}[!t]
   \centering
   \newcommand{\wid}{0.165\textwidth}
   \subfloat[Scale variations on \emph{doll}.]{%
   \includegraphics[width=\wid]{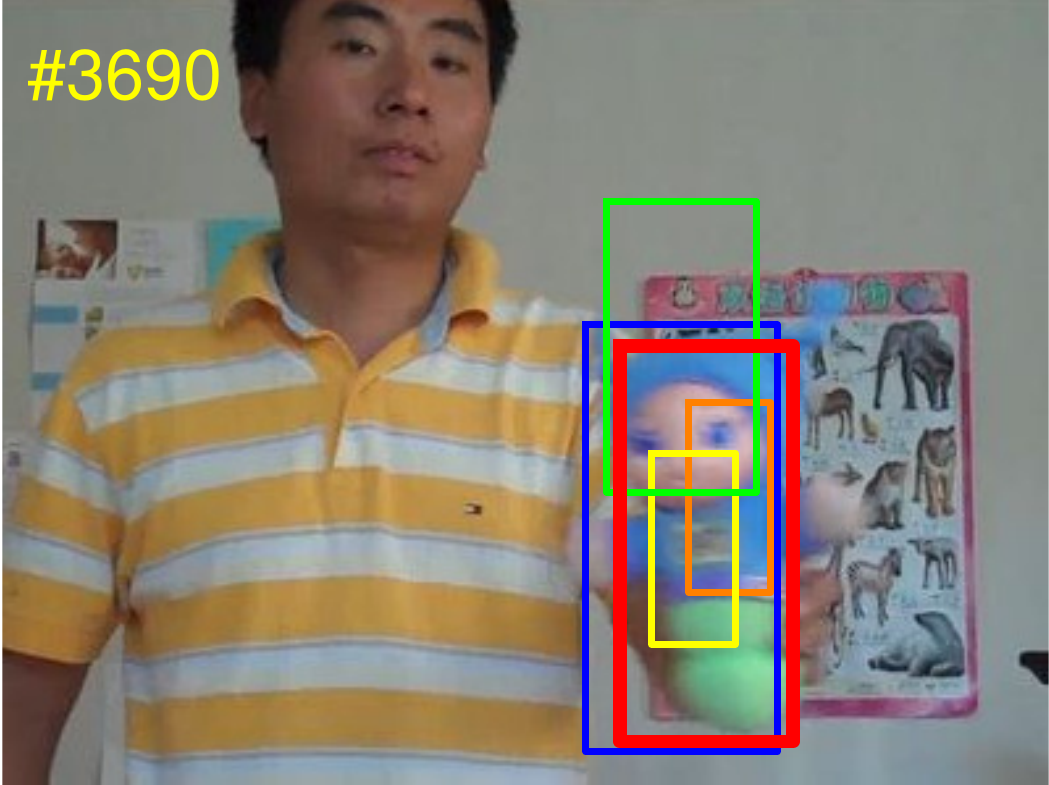}%
   \includegraphics[width=\wid]{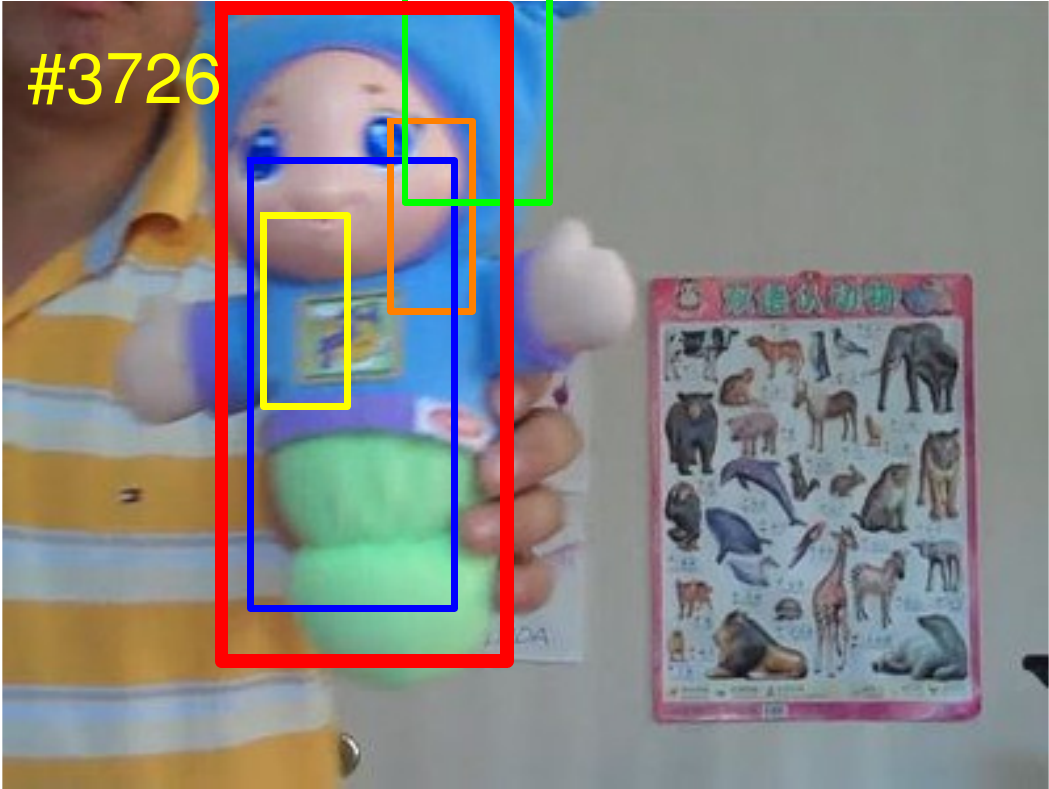}%
   \includegraphics[width=\wid]{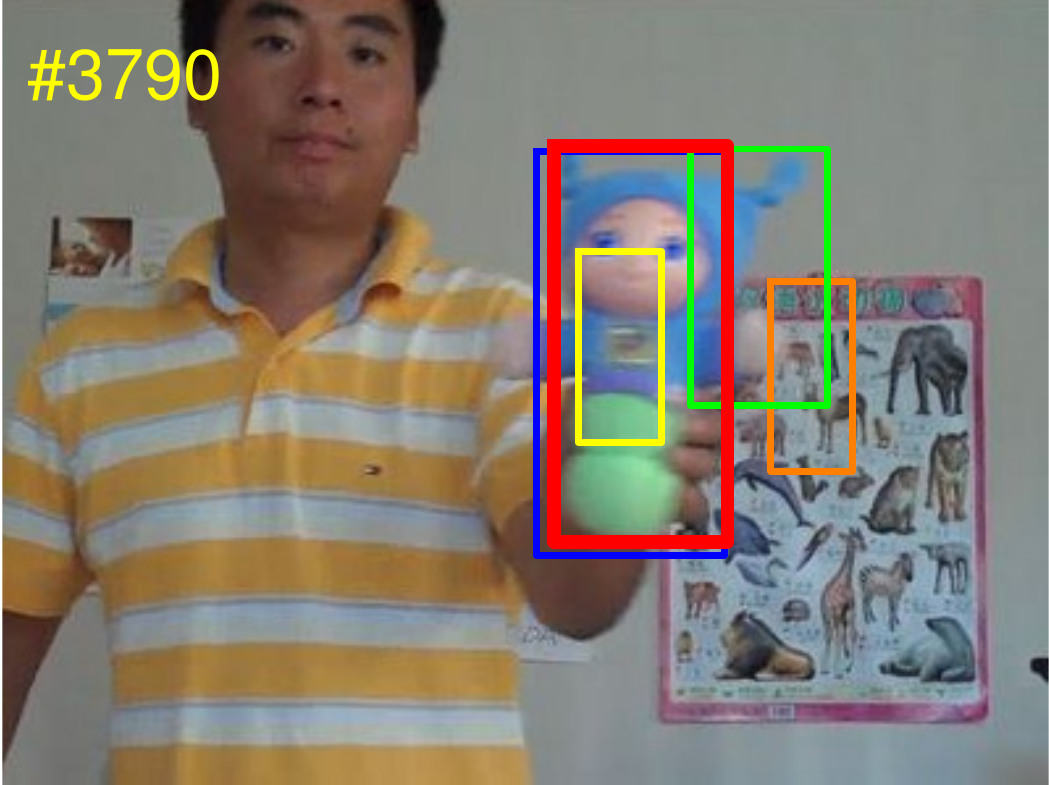}%
   \label{fig:scale_variation}}\hfil
   \subfloat[Out-of-plane rotations on \emph{trellis}.]{%
   \includegraphics[width=\wid]{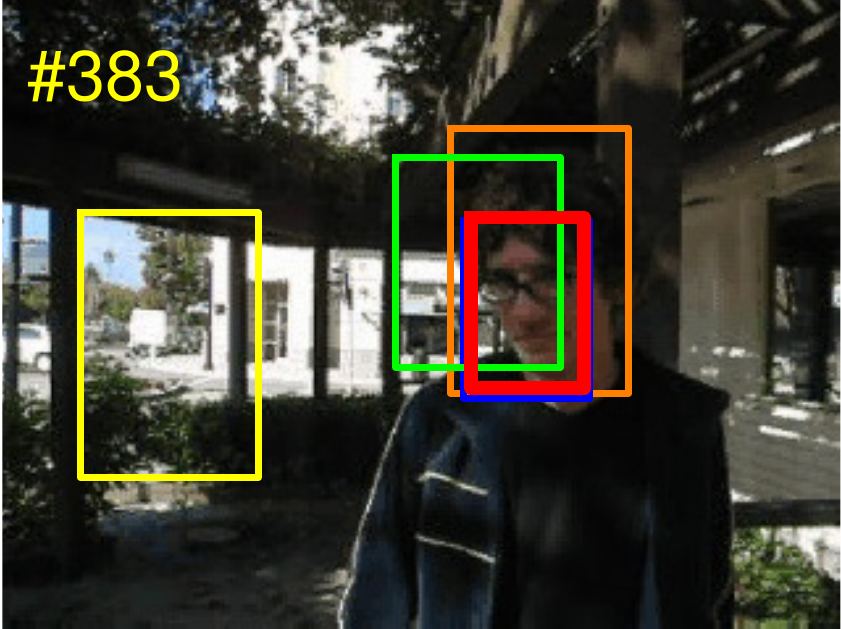}%
   \includegraphics[width=\wid]{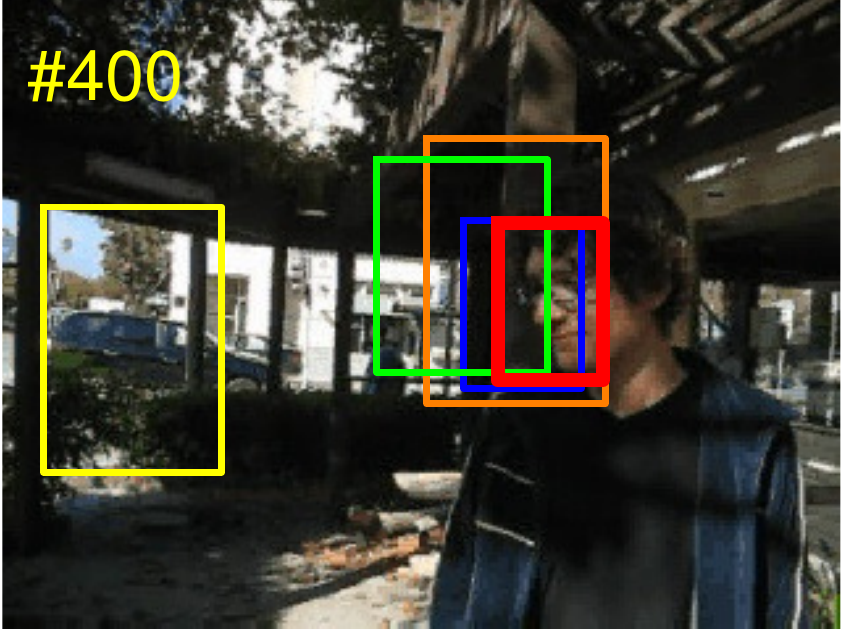}%
   \includegraphics[width=\wid]{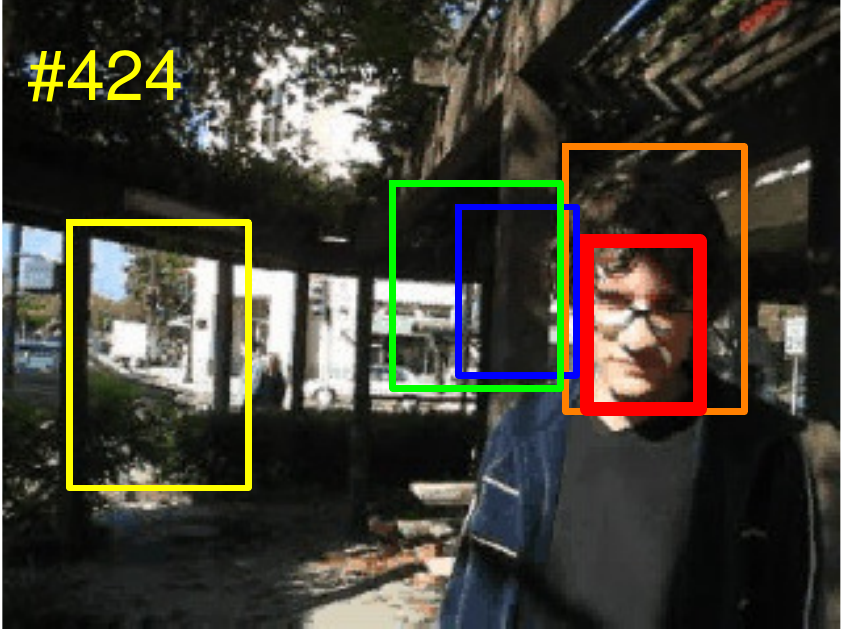}%
   \label{fig:rotation}}
   
   \subfloat[Illumination changes on \emph{skating}.]{%
   \includegraphics[width=\wid]{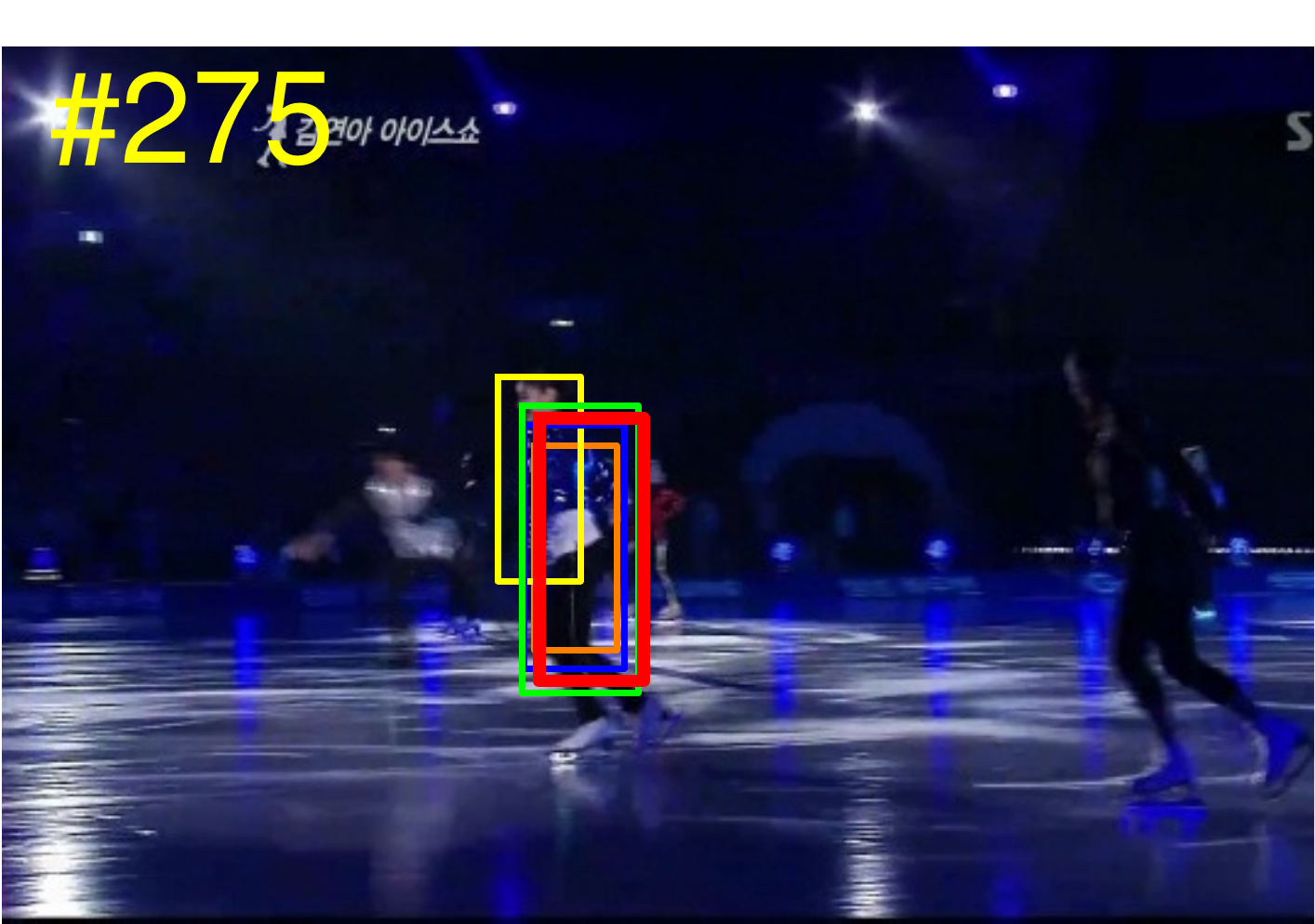}%
   \includegraphics[width=\wid]{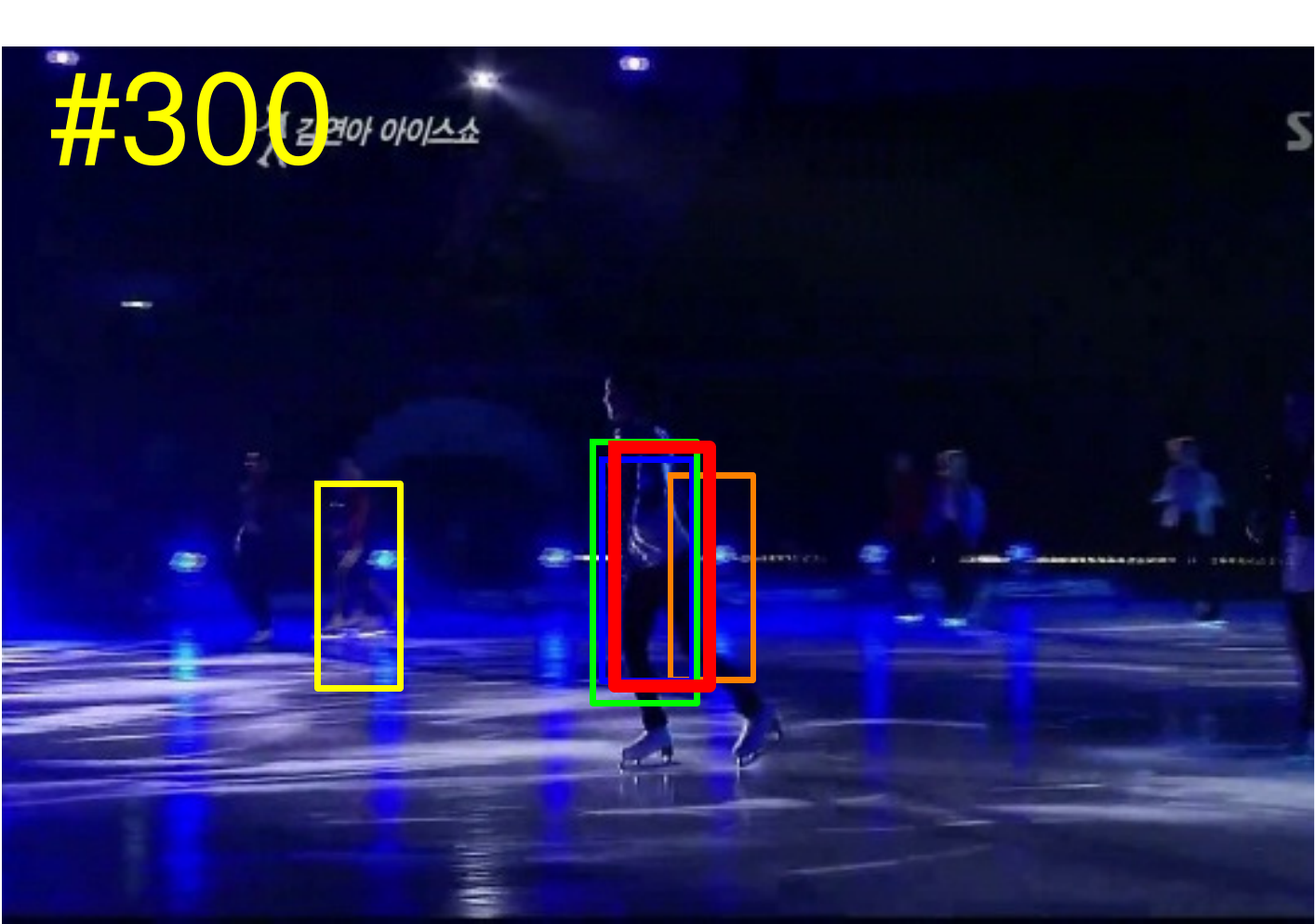}%
   \includegraphics[width=\wid]{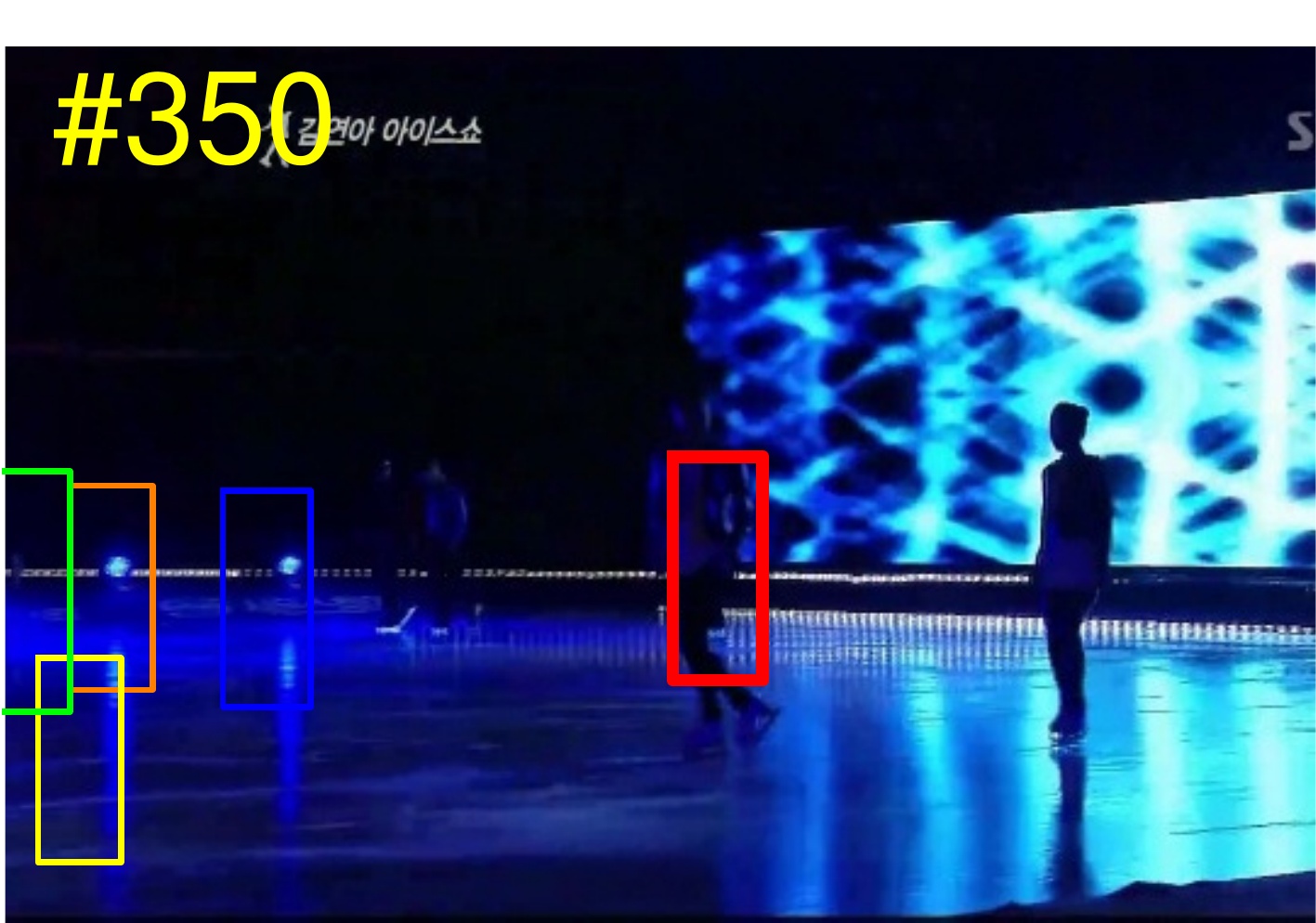}%
   \label{fig:illumination_variation}}\hfil
   \subfloat[Background clutter on \emph{soccer}.]{%
   \includegraphics[width=\wid]{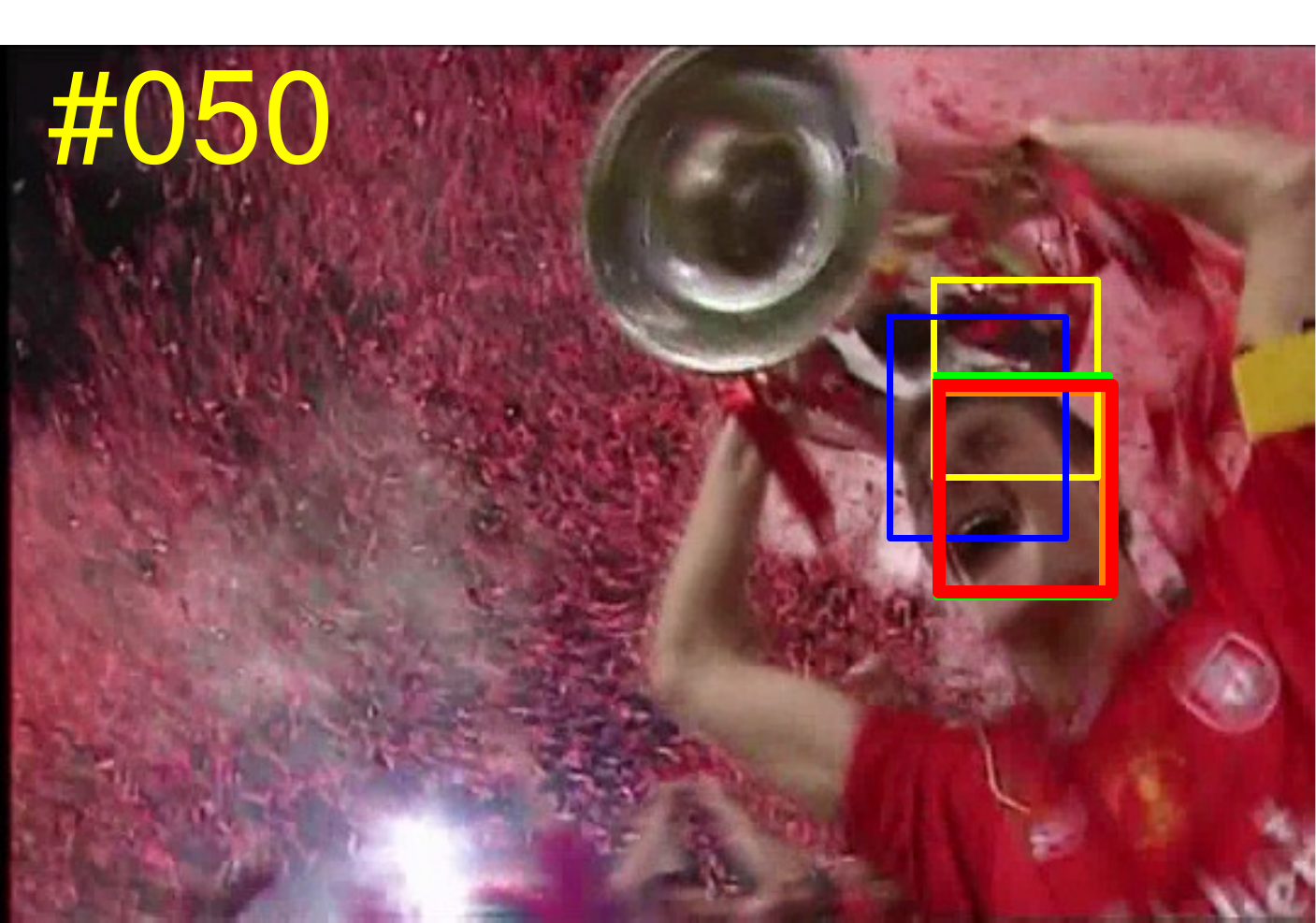}%
   \includegraphics[width=\wid]{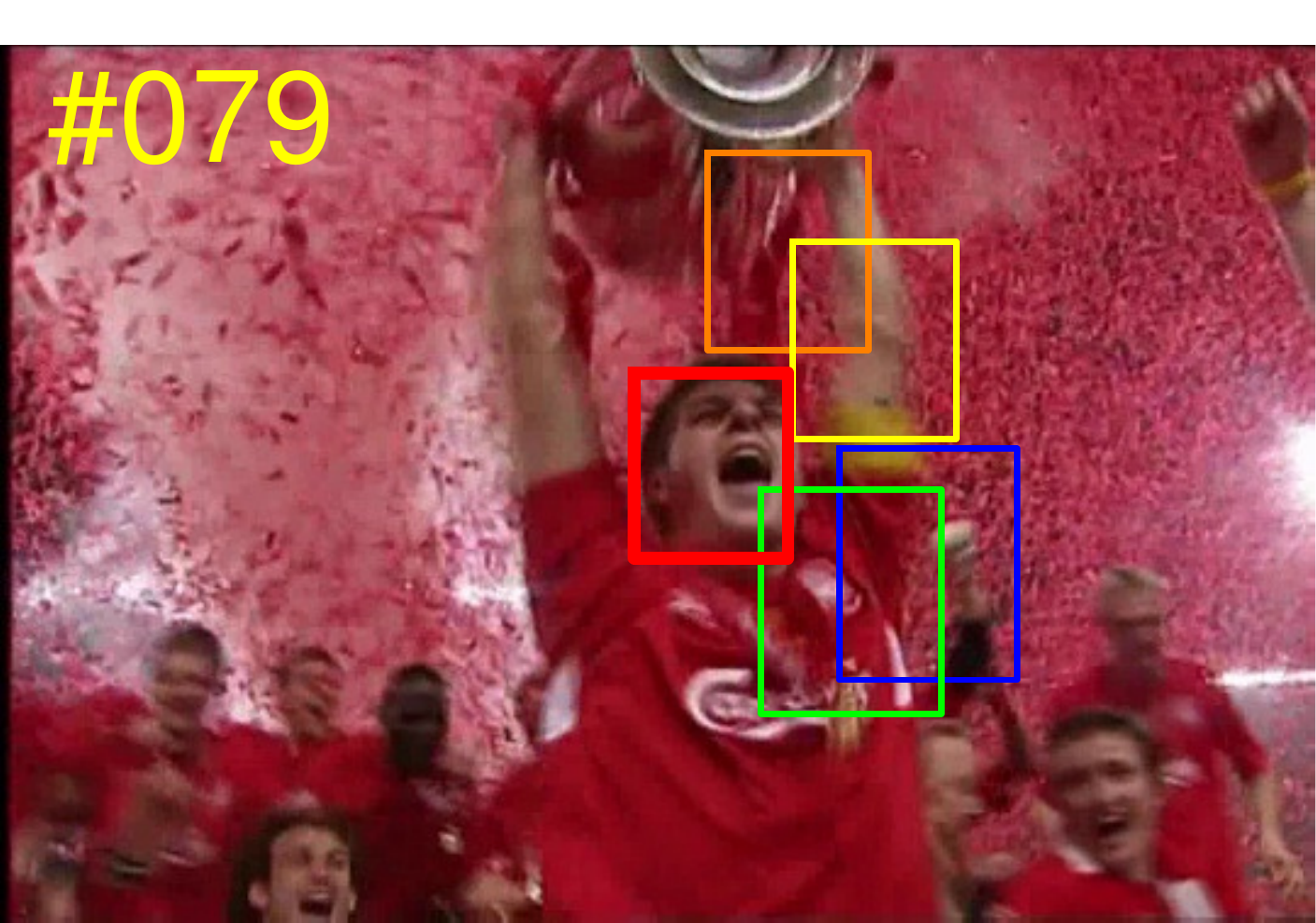}%
   \includegraphics[width=\wid]{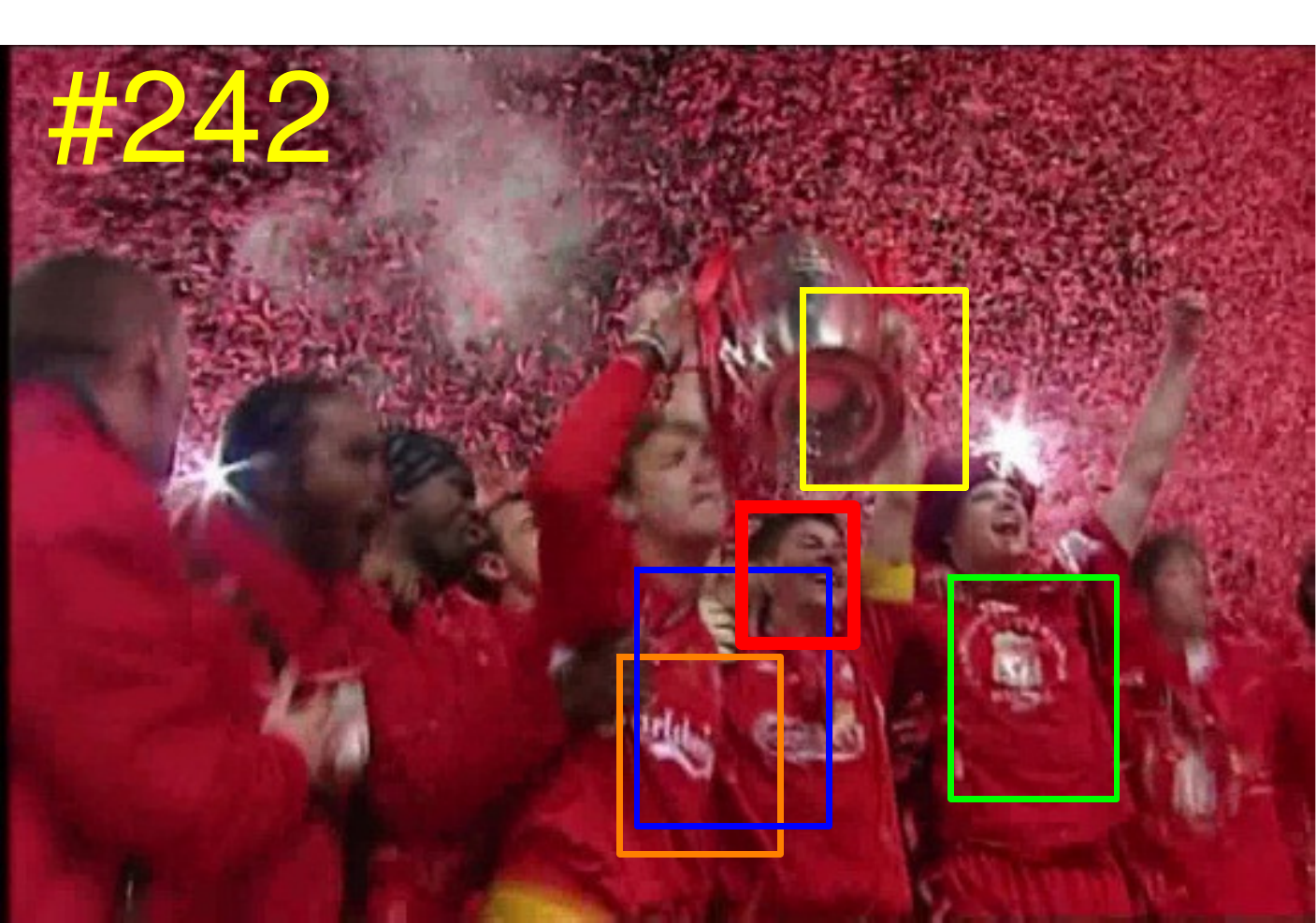}%
   \label{fig:clutter}}%
   \vspace{2mm}
   \includegraphics[width=0.5\textwidth, trim= 0 0 0 0]{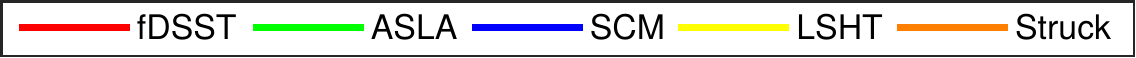}
  \caption{A qualitative comparison of our approach with four state-of-the-art trackers. Tracking results are shown on four example videos from the OTB dataset. The videos show challenging situations, such as scale variations \protect\subref{fig:scale_variation}, out-of-plane rotation \protect\subref{fig:rotation}, illumination variations \protect\subref{fig:illumination_variation} and partial occlusions \protect\subref{fig:clutter}. Our approach performs favorably compared to the existing tracker in these challenging situations.}%
  \label{fig:qual_eval}
\end{figure*}

\subsubsection{Attribute-based Comparison}
We also perform an attribute based analysis of our approach. In the OTB dataset, all videos are annotated with 11 different attributes, namely: in-plane rotation, scale variation, out of view, background clutter, illumination variation, motion blur, fast motion, deformation, out-of-plane rotation, occlusion and low resolution. 

Table~\ref{tab:attribute} contains the AUC scores for the 11 different attributes. For clarity, we report the results for the top ten trackers on the OTB dataset. Our approach provides favorable results on 7 out of 11 attributes: in-plane rotation, scale variation, background clutter, illumination variation, motion blur, fast motion and out-of-plane rotation. In sequences annotated with the scale variation attribute, our approach outperforms the compared correlation filter based translation-trackers (KCF and CSK). Moreover, our method outperforms the scale adaptive correlation based SAMF tracker by $5.1 \%$ in AUC. This shows that our tracker accurately estimates the size of the target and achieves superior performance in scenarios with scale variation.

\subsubsection{Comparison of Robustness to Initialization}
We also evaluate the robustness of our tracker with respect to initialization. The evaluation is performed as proposed in \cite{Wu13g}. Two different criteria, namely temporal robustness (TRE) and spatial robustness (SRE) are employed to evaluate the robustness of our approach. The SRE is performed by initializing the tracker at different locations near the ground-truth bounding box in the first frame. The tracker is evaluated on each video with 12 different initializations. Four of the perturbations are computed by shifting the box horizontally left and right, and vertically up and down. Another four are obtained by shifting the box in all four diagonal directions. The magnitude of these shifts are $10\%$ of the target size in the corresponding direction. The remaining four perturbations are obtained by only rescaling the ground-truth box with the factors $0.8$, $0.9$, $1.1$ and $1.2$. The TRE is performed by initializing the tracker at different frames with the ground-truth bounding box. In the TRE case, each video is partitioned into 20 segments. We refer to \cite{Wu13g} for more details.

Figure~\ref{fig:TRE} and \ref{fig:SRE} shows the success plots for the TRE and SRE analysis. For clarity, we compare with five top performing trackers. In both evaluations, our approach performs favorably compared to existing methods. It is worth mentioning that the standard SAMF employs multi-cue image representation (HOG and Color names). For a fair comparison, we also compare with \emph{SAMF HOG}, that employs the same feature representation as our fDSST. Our approach provides a consistent gain in performance compared to SAMF HOG for both TRE and SRE experiments. We expect further improvement in the performance of our tracker by incorporating the color information employed in SAMF. 

\subsubsection{Qualitative Evaluation}

Here we provide a qualitative comparison of our approach with existing trackers from the literature. Figure \ref{fig:qual_eval} illustrates frames from four sequences with illumination variation (skating), out-of-plane rotations (trellis), background clutter (soccer) and significant scale variations (doll). Among existing trackers, both ASLA and SCM are capable of estimating scale variations. In the doll sequence, both SCM and ASLA suffer from a significant scale drift in the presence of rotating motions and fast scale changes. Our approach accurately estimates the target scale and translation despite the mentioned factors. In the trellis sequence, the compared trackers struggle due to difficult lighting conditions and out-of-plane rotations, while our tracker robustly handles these factors. In the skating sequence, the multi-illuminant indoor lighting conditions together with the target deformations, cause most approaches to drift or fail. Again, our approach demonstrates robustness in these scenarios and is able to keep track of the target throughout the sequence. Finally, the compared trackers fail to handle the significant clutter and occlusions in the soccer sequence. In addition to robustly tracking the target, our approach accurately estimates the scale variations in this sequence.

\begin{figure*}[!t]
	\centering
	\newcommand{\wid}{0.36\textwidth}
	\includegraphics[width=\wid]{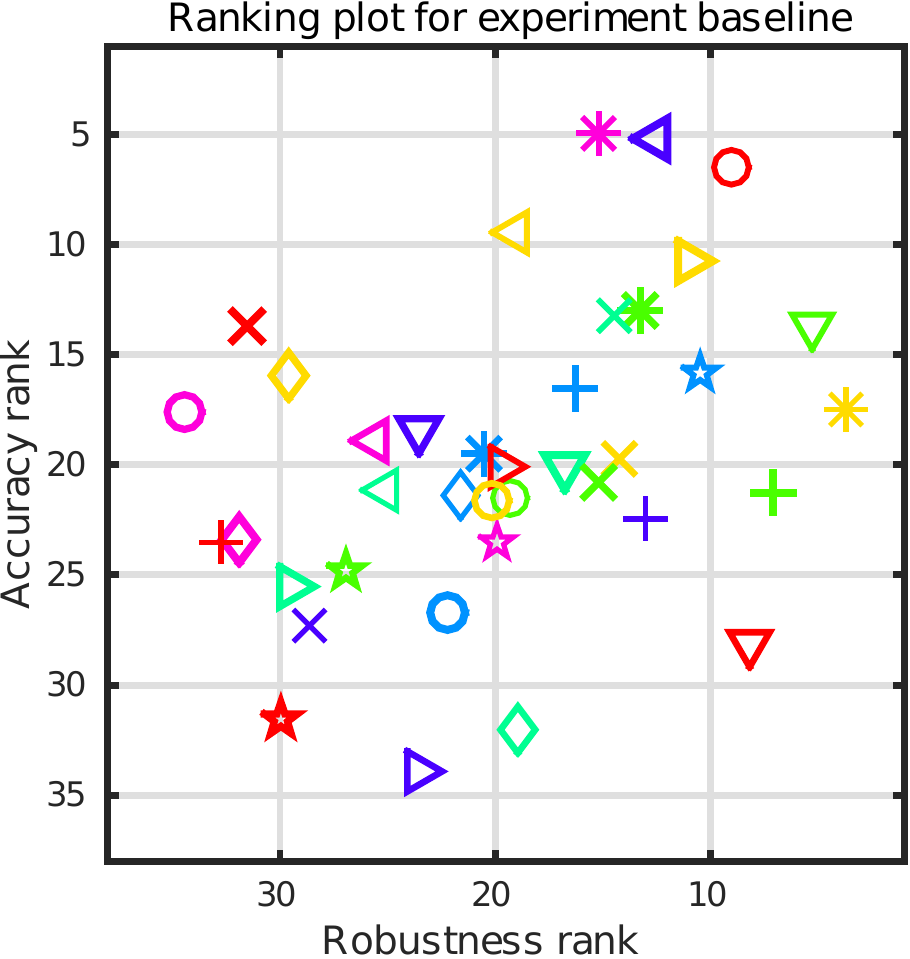}\hspace{7mm}
	\includegraphics[width=\wid]{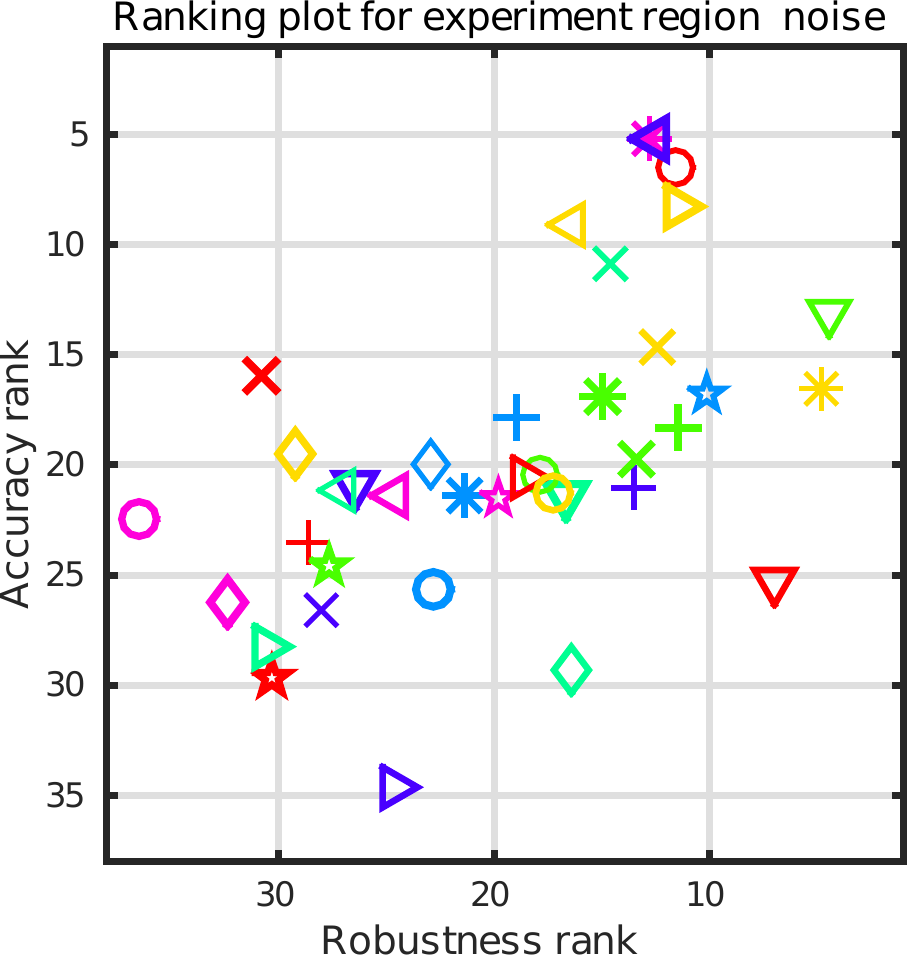}\vspace{2mm}
	\fbox{\includegraphics*[width=0.985\textwidth]{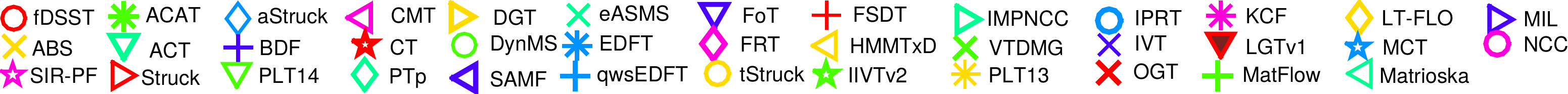}}
	\caption{Ranking plots for the baseline and region noise experiments in the VOT 2014 dataset. The accuracy and robustness rank are plotted along the vertical and horizontal axis respectively. Our fDSST approach (denoted by the red circle) achieves superior results in both experiments.}
	\label{fig:vot_rank}
\end{figure*}
\begin{table}[!t]
  \caption{A comparison of our fDSST with participating methods in the VOT 2014 challenge. The first four columns report the accuracy and robustness ranking scores in the baseline experiment and the region noise experiment. The final averaged ranking score is reported in the fifth column. Raw values for the average overlap and average number of failures are shown in the last two columns. Our method achieves the best final score.}
  \label{tab:vot_results}
  \centering
  \resizebox{\columnwidth}{!}{\begin{tabular}{l@{~~}c@{~~~}cc@{~~~}ccc@{~~~}c}
\toprule
& \multicolumn{2}{c}{\textbf{Baseline Experiment}} & \multicolumn{2}{c}{\textbf{Noise Experiment}} & \textbf{Final} & \multicolumn{2}{c}{\textbf{Raw Values}}\\ 
& Acc.\ Rank & Rob.\ Rank & Acc.\ Rank & Rob.\ Rank & Rank & Overlap & Failures \\\midrule
\textbf{fDSST}     & 6.44  & 9.06  & 6.48  & 11.54 & \textbf{\textcolor{red}{8.38}} & 0.61 & 1.04 \\
SAMF      & 5.22   & 12.56 & 5.24  & 12.54 & \textit{\textcolor{blue}{8.89}}  & 0.61 & 1.28 \\
PLT14     & 13.71  & 5.33  & 13.11 & 4.50  & 9.16  & 0.56 & 0.16 \\
KCF       & 4.96   & 15.17 & 5.17  & 12.75 & 9.51  & 0.62 & 1.32 \\
DGT       & 10.78  & 11.00 & 8.24  & 11.50 & 10.38 & 0.58 & 1.00 \\
PLT13     & 17.54  & 3.75  & 16.49 & 4.75  & 10.63 & 0.55 & 0.08 \\
eASMS     & 13.23  & 14.50 & 10.88 & 14.62 & 13.31 & 0.55 & 1.12 \\
MCT       & 15.88  & 10.54 & 16.75 & 10.17 & 13.33 & 0.53 & 0.99 \\
HMMTxD    & 9.43   & 19.08 & 9.12  & 16.39 & 13.51 & 0.59 & 1.52 \\
ACAT      & 12.99  & 13.23 & 16.90 & 14.99 & 14.53 & 0.55 & 1.56 \\
MatFlow   & 21.25  & 7.17  & 18.33 & 11.42 & 14.54 & 0.49 & 0.76 \\
ABS       & 19.72  & 14.25 & 14.69 & 12.42 & 15.27 & 0.52 & 1.24 \\
LGTv1     & 28.12  & 8.17  & 25.25 & 7.01  & 17.14 & 0.46 & 0.66 \\
VTDMG     & 20.77  & 15.25 & 19.72 & 13.42 & 17.29 & 0.52 & 1.32 \\
qwsEDFT   & 16.57  & 16.33 & 17.90 & 18.98 & 17.45 & 0.54 & 1.36 \\
BDF       & 22.42  & 13.08 & 20.99 & 13.54 & 17.51 & 0.49 & 1.20 \\
ACT       & 20.08  & 16.83 & 21.36 & 16.68 & 18.74 & 0.53 & 1.48 \\
Struck    & 20.11  & 19.72 & 20.60 & 18.62 & 19.76 & 0.51 & 2.16 \\
DynMS     & 21.54  & 19.31 & 20.42 & 17.92 & 19.80 & 0.51 & 1.54 \\
tStruck   & 21.63  & 20.21 & 21.26 & 17.26 & 20.09 & 0.50 & 2.22 \\
EDFT      & 19.51  & 20.56 & 21.39 & 21.42 & 20.72 & 0.52 & 1.84 \\
SIR-PF    & 23.54  & 19.92 & 21.49 & 19.83 & 21.20 & 0.49 & 1.94 \\
aStruck   & 21.41  & 21.58 & 19.98 & 23.00 & 21.49 & 0.50 & 2.44 \\
FoT       & 18.48  & 23.58 & 20.97 & 26.42 & 22.36 & 0.51 & 2.28 \\
CMT       & 18.93  & 25.58 & 21.37 & 24.58 & 22.61 & 0.48 & 2.64 \\
OGT       & 13.76  & 31.50 & 15.91 & 30.78 & 22.99 & 0.54 & 3.34 \\
LT-FLO    & 15.98  & 29.58 & 19.50 & 29.25 & 23.58 & 0.54 & 2.56 \\
Matrioska & 21.15  & 25.14 & 21.19 & 27.00 & 23.62 & 0.49 & 2.48 \\
PTp 	  & 32.05  & 18.92 & 29.26 & 16.44 & 24.17 & 0.44 & 1.40 \\
IPRT 	  & 26.68  & 22.25 & 25.62 & 22.77 & 24.33 & 0.47 & 1.86 \\
IIVTv2 	  & 24.79  & 26.92 & 24.58 & 27.72 & 26.00 & 0.47 & 3.19 \\
FSDT 	  & 23.55  & 32.75 & 23.58 & 28.66 & 27.14 & 0.46 & 3.08 \\
IVT 	  & 27.31  & 28.67 & 26.60 & 28.00 & 27.65 & 0.47 & 2.76 \\
NCC 	  & 17.65  & 34.42 & 22.43 & 36.50 & 27.75 & 0.47 & 7.64 \\
IMPNCC 	  & 25.56  & 29.42 & 28.29 & 30.58 & 28.46 & 0.47 & 3.64 \\
FRT 	  & 23.38  & 31.88 & 26.21 & 32.42 & 28.47 & 0.48 & 3.32 \\
MIL 	  & 33.95  & 23.58 & 34.61 & 24.60 & 29.19 & 0.39 & 2.27 \\
CT        & 31.51  & 29.92 & 29.66 & 30.33 & 30.36 & 0.43 & 3.12 \\
\bottomrule
\end{tabular}

}\vspace{0mm}
\end{table}

\begin{table*}[!t]
	\caption{Attribute analysis on the VOT 2014 dataset. For each of the five attributes, we report the average overlap and the number of tracking failures. We also report the results for frames with no annotated attribute. For clarity, we only show the results for the top ten trackers in the VOT 2014 challenge. The best and second best entries for each attribute are shown in red and blue font respectively.}
	\label{tab:vot_attribute}
	\centering
	\resizebox{\textwidth}{!}{\newcommand{\first}[1]{\textbf{\textcolor{red}{#1}}}%
\newcommand{\second}[1]{\textit{\textcolor{blue}{#1}}}%
\begin{tabular}{lc@{~~~}cc@{~~~}cc@{~~~}cc@{~~~}cc@{~~~}cc@{~~~}c}
	\toprule
	& \multicolumn{2}{c}{\textbf{Camera motion}} & \multicolumn{2}{c}{\textbf{Illumination change}} & \multicolumn{2}{c}{\textbf{Occlusion}} & \multicolumn{2}{c}{\textbf{Size change}} & \multicolumn{2}{c}{\textbf{Motion change}} & \multicolumn{2}{c}{\textbf{No attribute label}}\\
	& Failures & Overlap & Failures & Overlap & Failures & Overlap & Failures & Overlap & Failures & Overlap & Failures & Overlap\\\midrule
	\textbf{fDSST} & 19.0 & 0.65 & \first{0.0} & \second{0.72} & 4.0 & \second{0.61} & 12.0 & 0.53 & 23.0 & \second{0.64} & \first{0.0} & 0.57\\
	SAMF & 24.0 & \second{0.66} & \second{1.0} & 0.67 & 4.0 & \second{0.61} & 18.0 & \second{0.56} & 25.0 & \first{0.67} & \first{0.0} & 0.57\\
	PLT-14 & \second{4.0} & 0.56 & \second{1.0} & 0.5 & \second{2.0} & 0.59 & \second{4.0} & 0.51 & \second{4.0} & 0.57 & \first{0.0} & 0.53\\
	KCF & 24.0 & \first{0.67} & \second{1.0} & \first{0.74} & 5.0 & \first{0.64} & 20.0 & \first{0.58} & 26.0 & \first{0.67} & \first{0.0} & 0.54\\
	DGT & 19.0 & 0.56 & 14.0 & 0.47 & \first{1.0} & 0.48 & 6.0 & \first{0.58} & 14.0 & 0.58 & \first{0.0} & \first{0.68}\\
	PLT-13 & \first{2.0} & 0.55 & \first{0.0} & 0.52 & \first{1.0} & 0.58 & \first{2.0} & 0.48 & \first{2.0} & 0.55 & \first{0.0} & 0.49\\
	eASMS & 25.0 & 0.55 & 14.0 & 0.46 & 5.0 & 0.56 & 6.0 & 0.50 & 10.0 & 0.55 & \first{0.0} & 0.59\\
	MCT & 15.47 & 0.55 & 1.2 & 0.57 & 2.27 & 0.5 & 12.4 & 0.48 & 20.07 & 0.54 & \second{0.27} & 0.55\\
	HMMTxD & 32.0 & 0.6 & 10.0 & 0.58 & 7.0 & 0.59 & 18.0 & 0.54 & 20.0 & 0.61 & \first{0.0} & 0.55\\
	ACAT & 28.0 & 0.55 & \second{1.0} & 0.62 & 4.0 & 0.49 & 16.0 & 0.49 & 29.0 & 0.57 & \first{0.0} & \second{0.62}\\
	\bottomrule
\end{tabular}

}
	\vspace{0mm}
\end{table*}

\subsection{Experiment 4: VOT Challenge 2014}
\label{sec:vot2014}
Here we present the results on the VOT 2014 dataset \cite{VOT2014}. We compare our fDSST approach with 37 participating trackers in the challenge. In addition to the baseline experiment, we evaluate the robustness by performing the \emph{region noise} experiment in the VOT protocol. This is performed by introducing noise in the initial bounding boxes and evaluating the tracker multiple times for each video. For more details about this experiment, we refer to \cite{VOT2014}.

Table~\ref{tab:vot_results} presents the final ranking score over all the 25 videos in the VOT 2014 dataset. In addition to the final average rank, the partial results for the baseline and region noise experiments are shown. We also report the average overlap and failure rate over all videos in the baseline experiment. Our fDSST approach achieves the top average rank among all the 38 trackers. Figure~\ref{fig:vot_rank} shows the ranking plots for the baseline and region noise experiments. Our approach performs favorably both in terms of accuracy and robustness compared to the other trackers. Contrary to the OTB dataset, Struck provides inferior performance on this dataset. The second best method, SAMF \cite{Li2014}, is based on a kernelized translation filter and applies a multi-resolution strategy for estimating the target scale. While our fDSST only uses HOG and grayscale features, SAMF employs multiple features for tracking. In particular, it uses color information by combining HOG and intensity with the Color Names representation \cite{Weijer09a}.

In the VOT 2014 dataset, the videos are annotated by five different attributes: Camera motion, illumination change, occlusion, size change and motion change. Different from the OTB dataset, the attributes in VOT are annotated per-frame in a video. Table~\ref{tab:vot_attribute} shows the average overlap and failures for all attributes on VOT 2014. Only results for the top ten trackers in the VOT challenge are reported for clarity. 

Compared to the second best approach (SAMF), our method obtains a reduced failure rate for four attributes. On the other hand, the PLT-13 and PLT-14 achieves better robustness on all attributes compared to the correlation based trackers (fDSST, SAMF and KCF). Similar to Struck \cite{Torr11b}, the PLT methods are also based on an online structural SVM detector. In addition, the PLT approaches employ a feature selection strategy, leading to superior robustness. Note that our fDSST does not employ any feature selection. Unlike PLT, correlation based trackers compute a dense set of classification scores in a limited search region. This likely explains the superior accuracy of correlation trackers, such as fDSST, compared to PLT.

The success of our approach on both datasets clearly suggests the importance of accurate scale estimation for visual tracking. It is worth mentioning that our approach employs exactly the same parameter settings for both datasets.

\section{Conclusion}
\label{sec:conclusion}
In this paper, we investigate the problem of accurate and robust scale estimation for real-time visual tracking. We propose a novel scale-adaptive approach for accurately estimating the size of the target. Our approach is based on learning separate discriminative correlation filters for translation and scale estimation. The explicit scale filter is directly learned from samples of the appearance change induced by scale variations. Furthermore, we propose strategies to reduce the computational cost of our tracking approach. This allows us to use a larger target search space without sacrificing real-time performance. We show that this results in a significant increase in tracking performance and a twofold gain in speed.

We conduct extensive experiments on the Online Tracking Benchmark (OTB) and the Visual Object Tracking (VOT) 2014 challenge datasets. The results clearly demonstrate that our approach provides significant improvement over the baseline translation-tracker. We also compare our approach with several state-of-the-art trackers. Our method outperforms 19 state-of-the-art trackers in the literature on the OTB dataset. On the VOT 2014 dataset, our tracking approach is shown to outperform 37 state-of-the-art trackers, obtaining the top combined ranking score.

In this work, we only employ intensity based image representation. Future work includes investigating efficient feature fusion strategies to combine intensity and color information. Incorporating color information is expected to improve the performance of our tracker, especially in scenarios with target deformations. Another research direction is to investigate the applicability of our tracker in visual surveillance scenarios.

\appendix  

%

Here, we derive of the solution \eqref{eq:multiple_feat_sol} of the filter that minimizes the loss \eqref{eq:multiple_feat_cost}. Applying Parseval's formula to \eqref{eq:multiple_feat_cost} and exploiting the correlation property of the DFT gives,
\begin{align}
	\label{eq:multiple_Feat_cost_dft}
	\tilde{\varepsilon} &= \left\|G - \sum_{l=1}^{d} \overline{H^l} F^l \right\|^2 + \lambda \sum_{l=1}^{d} \left\| H^l \right\|^2  \\
	&= \sum_{\bm{n}} \left(\left|G(\bm{n}) - \sum_{l=1}^{d} \overline{H^l(\bm{n})} F^l(\bm{n}) \right|^2 + \lambda \sum_{l=1}^{d} \left| H^l(\bm{n}) \right|^2\right) \nonumber  \\
	&= \sum_{\bm{n}} \left(\left|\overline{G(\bm{n})} - F(\bm{n})\ctp H(\bm{n}) \right|^2 + \lambda \left\| H(\bm{n}) \right\|^2\right) . \nonumber
\end{align}
Here, the sum over $\bm{n}$ ranges over all discrete frequencies in the DFT. Note that $F(\bm{n}), H(\bm{n}) \in \complexes^d$ are $d$-dimensional complex vectors and $F(\bm{n})\ctp H(\bm{n})$ is their inner product. We use $\ctp$ to denote the conjugate transpose of a matrix.

Each term $\bm{n}$ in \eqref{eq:multiple_Feat_cost_dft} can be minimized independently since it only depends on the DFT coefficients $H(\bm{n})$ at frequency $\bm{n}$. We thus get a separate quadratic minimization problem for each frequency $\bm{n}$. The solution is given by solving the resulting normal equations derived from each term in \eqref{eq:multiple_Feat_cost_dft},
\begin{equation}
	\label{eq:multiple_Feat_normaleq}
	\left( F(\bm{n}) F(\bm{n})\ctp + \lambda I_d\right) H(\bm{n}) = F(\bm{n}) \overline{G(\bm{n})} .
\end{equation}
Here, $I_d$ denotes the $d\times d$ identity matrix. The normal equations \eqref{eq:multiple_Feat_normaleq} can be solved analytically using the formula for the inverse of a rank-1 adjustment \cite{HornJohnson}: $(x y\ctp + A)^{-1} = A^{-1} - (y\ctp A^{-1} x + 1)^{-1} A^{-1} x y\ctp A^{-1}$. Here, $A$ is a nonsingular $m \times m$ matrix, while $x$ and $y$ are $m$-dimensional vectors. Using $A = \lambda I_d$ and $x = y = F(\bm{n})$ we obtain,
\begin{align}
	\label{eq:multiple_Feat_derivation}
	H(\bm{n}) &= \left( F(\bm{n}) F(\bm{n})\ctp + \lambda I_d\right)^{-1} F(\bm{n}) \overline{G(\bm{n})} \nonumber \\
	&= \left( \frac{1}{\lambda} I_d - \frac{1}{\lambda} \frac{F(\bm{n}) F(\bm{n})\ctp}{F(\bm{n})\ctp F(\bm{n}) + \lambda} \right) F(\bm{n}) \overline{G(\bm{n})} \nonumber \\
	&= \left( F(\bm{n}) - \frac{F(\bm{n}) F(\bm{n})\ctp F(\bm{n})}{F(\bm{n})\ctp F(\bm{n}) + \lambda} \right) \frac{\overline{G(\bm{n})}}{\lambda} \nonumber \\
	&= F(\bm{n}) \left( 1 - \frac{F(\bm{n})\ctp F(\bm{n})}{F(\bm{n})\ctp F(\bm{n}) + \lambda} \right) \frac{\overline{G(\bm{n})}}{\lambda} \nonumber \\
	&= \frac{\overline{G(\bm{n})} }{F(\bm{n})\ctp F(\bm{n}) + \lambda} F(\bm{n}) .
\end{align}
This expression is equivalent to \eqref{eq:multiple_feat_sol}.

\ifCLASSOPTIONcompsoc
  \section*{Acknowledgments}
\else
  \section*{Acknowledgment}
\fi
%
%

This work has been supported by the Swedish Foundation for Strategic Research through a grant for the project CUAS, by the Swedish Research Council through a grant for the project EMC${}^2$, by the Strategic Vehicle Research and Innovation (FFI) through a grand for the project CCAIS, by the Wallenberg Autonomous Systems Program, by the National Supercomputer Centre and Nvidia.

\ifCLASSOPTIONcaptionsoff
  \newpage
\fi



\bibliographystyle{IEEEtran}
\bibliography{scaletracking_journal}

%








\vfill

\newpage

\begin{IEEEbiography}[{\includegraphics[width=1in,height=1.25in,clip,keepaspectratio]{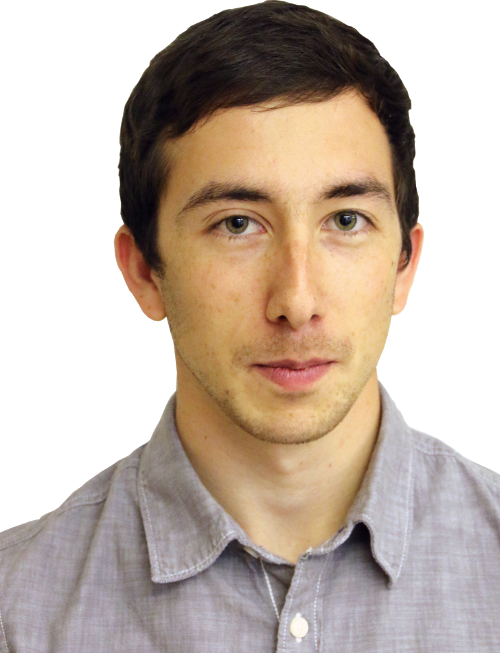}}]{Martin Danelljan}
	is a Ph.D.\ student at the Computer Vision Laboratory, Link\"oping University, Sweden. He received his M.Sc.\ degree in Electrical Engineering from Link\"oping University in 2014. His research interests include machine learning methods for visual tracking and statistical models for point set registration. He has published several papers in major computer vision conferences, including CVPR, ICCV and ECCV. In 2014, he was awarded the \textit{Tryggve Holm} medal for outstanding student achievements and grades and the Swedish Computer Society award for best master's thesis. He has achieved the top rank in the Visual Object Tracking (VOT) Challenge 2014, the OpenCV State-of-the-Art Vision Challenge 2015 in Tracking and the VOT Thermal Infrared Challenge 2015.
\end{IEEEbiography}

\begin{IEEEbiography}[{\includegraphics[width=1in,height=1.25in,clip,keepaspectratio]{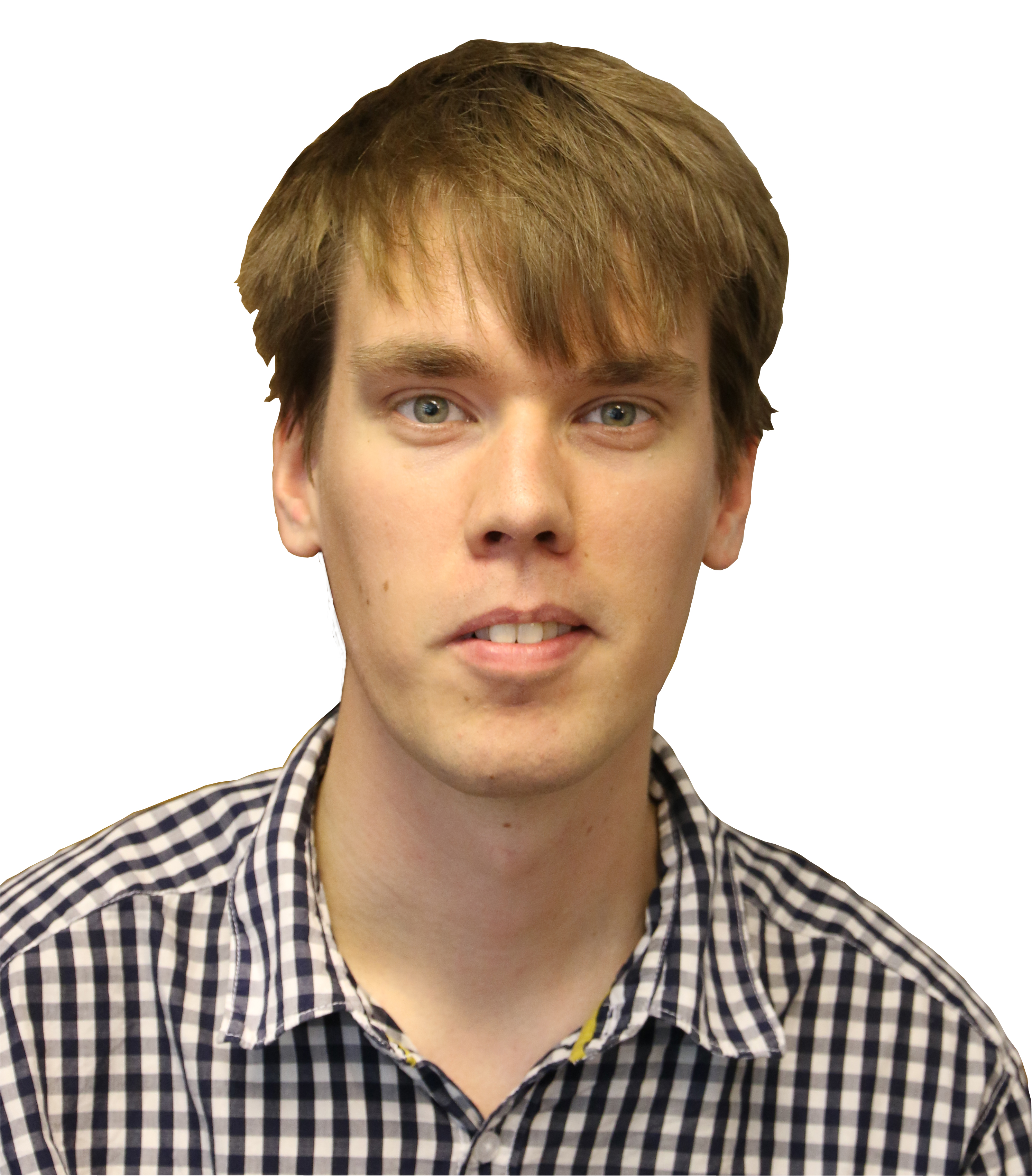}}]{Gustav H\"ager}
	is a Ph.D.\ student at Link\"oping University. His research interests include machine learning, with particular focus on methods for online tracking and detection. He also supervises the LiU-humanoids standard platform league team for the RoboCup competitions.
\end{IEEEbiography}

\begin{IEEEbiography}[{\includegraphics[width=1in,height=1.25in,clip,keepaspectratio]{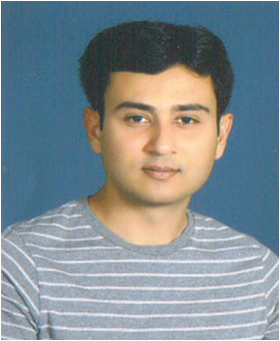}}]{Fahad Shahbaz Khan}
	is a research fellow at Computer Vision Laboratory, Link\"oping University, Sweden. He received the M.Sc.\ degree in Intelligent Systems Design from Chalmers University of Technology, Sweden and a Ph.D.\ degree in Computer Vision from Autonomous University of Barcelona, Spain. From 2012 to 2014, he was post doctoral fellow at Computer Vision Laboratory, Link\"oping University, Sweden. His research interests are in object recognition, action recognition and visual tracking. He has published articles in high-impact computer vision journals and conferences in these areas.
\end{IEEEbiography}

\begin{IEEEbiography}[{\includegraphics[width=1in,height=1.25in,clip,keepaspectratio]{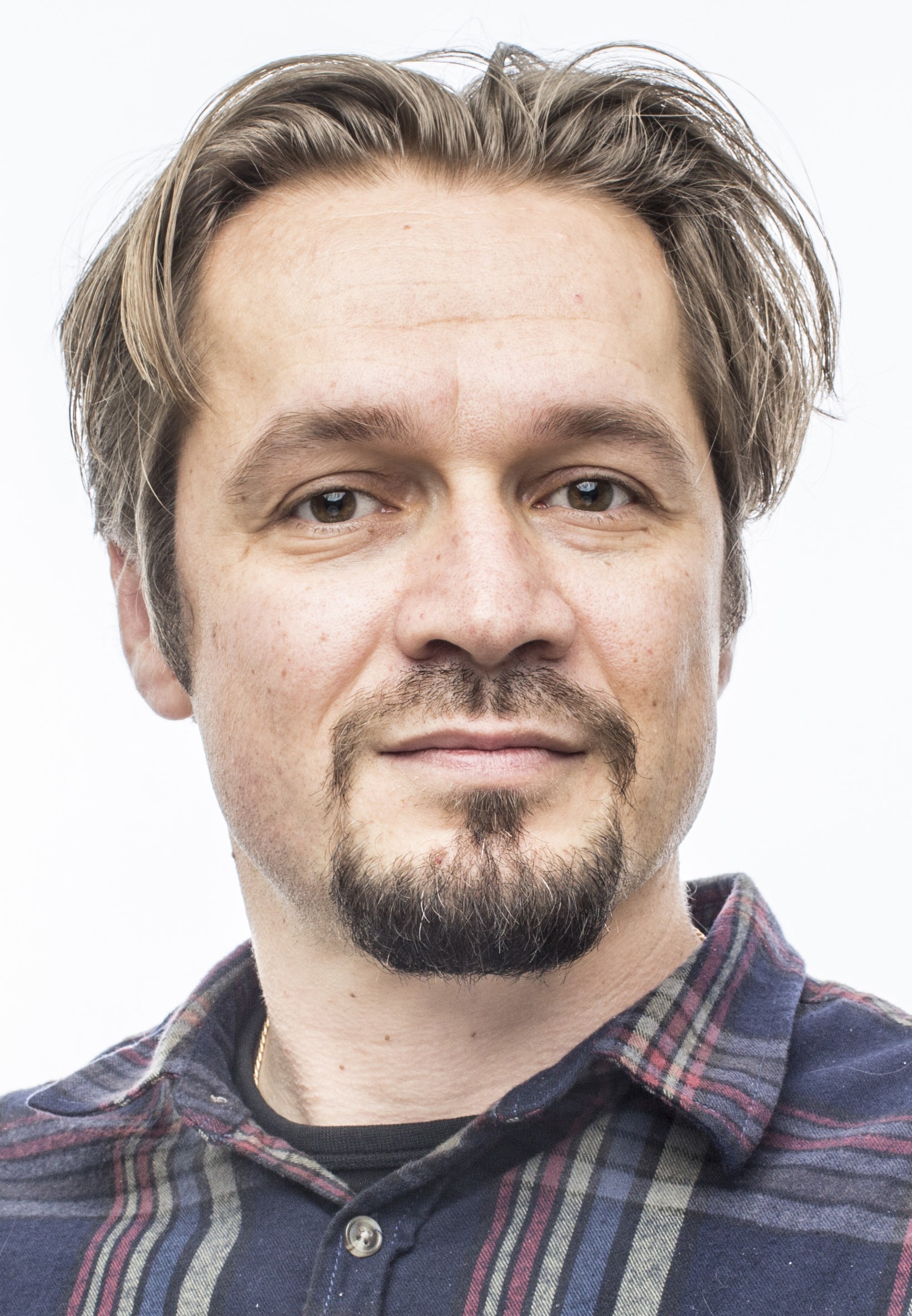}}]{Michael Felsberg}
	Michael Felsberg received the Ph.D. degree in engineering from the University of Kiel, Kiel, Germany, in 2002. Since 2008, he has been a Full Professor and the Head of the Computer Vision Laboratory, Link\"oping University, Link\"oping, Sweden. His current research interests include signal processing methods for image analysis, computer and robot vision, and machine learning. He has published more than 100 reviewed conference papers, journal articles, and book contributions. He was a recipient of awards from the German Pattern Recognition Society in 2000, 2004, and 2005, from the Swedish Society for Automated Image Analysis in 2007 and 2010, from Conference on Information Fusion in 2011 (Honorable Mention), and from the CVPR Workshop on Mobile Vision 2014. He has achieved top ranks on various challenges (VOT: 3rd 2013, 1st 2014, 2nd 2015; VOT-TIR: 1st 2015; OpenCV Tracking: 1st 2015; KITTI Stereo Odometry: 1st 2015, March). He has coordinated the EU projects COSPAL and DIPLECS, he is an Associate Editor of the Journal of Mathematical Imaging and Vision, Journal of Image and Vision Computing, Journal of Real-Time Image Processing, Frontiers in Robotics and AI. He was Publication Chair of the International Conference on Pattern Recognition 2014 and Track Chair 2016, he was the General Co-Chair of the DAGM symposium in 2011, and he will be general Chair of CAIP 2017.
\end{IEEEbiography}

\vfill

\end{document}